%% file: template.tex
\documentclass{article}

\usepackage{arxiv}

\usepackage[utf8]{inputenc} % allow utf-8 input
\usepackage[T1]{fontenc}    % use 8-bit T1 fonts
\usepackage[table,dvipsnames]{xcolor}
\usepackage{url}            % simple URL typesetting
\usepackage{booktabs}       % professional-quality tables
\usepackage{amsfonts}       % blackboard math symbols
\usepackage{amsmath,amssymb}
\usepackage{microtype}      % microtypography
\usepackage{graphicx}
\usepackage{multirow}
\usepackage{subcaption}     % subfigures
\usepackage{natbib}         % author--year and numerical citations
\usepackage{hyperref}

\definecolor{Reza}{RGB}{0, 100, 255}
\newif\ifshowtodos \showtodostrue

\newcommand{\scat}[1]{\includegraphics[width=0.232\textwidth]{plots/scatter/#1.pdf}}
\newcommand{\rowlab}[1]{\raisebox{0.105\textwidth}{\rotatebox{90}{\makebox[0pt][c]{\small #1}}}}

\title{Climbing the Design Ladder: Sequential Knowledge Distillation for Early-Stage Circuit Timing Prediction}

\renewcommand{\shorttitle}{Climbing the Design Ladder}
\date{October 2026}

\hypersetup{
  pdftitle={Climbing the Design Ladder: Sequential Knowledge Distillation for Early-Stage Circuit Timing Prediction},
  pdfsubject={cs.LG, cs.AR},
  pdfauthor={Reza Moravej, Fahad Rahman Amik, Zhanguang Zhang, Didier Ch\'etelat, Yingxue Zhang},
  pdfkeywords={timing prediction, knowledge distillation, electronic design automation, graph neural networks},
}

\author{%
  Reza Moravej \\
  Huawei Noah's Ark Lab \\
  \texttt{reza.moravej@huawei.com}
  \And
  Fahad Rahman Amik \thanks{Work done during internship at Huawei}\\
  McGill University \\
  \texttt{fahad.amik@mail.mcgill.ca}
  \And
  Zhanguang Zhang \\
  Huawei Noah's Ark Lab \\
  \texttt{zhanguang.zhang@huawei.com}
  \And
  Didier Chételat \\
  Huawei Noah's Ark Lab \\
  \texttt{didier.chetelat@huawei.com}
  \And
  Yingxue Zhang \\
  Huawei Noah's Ark Lab \\
  \texttt{yingxue.zhang@huawei.com}
}

\begin{document}

\maketitle

\begin{abstract}
Integrated circuit design involves multiple design stages: logic synthesis, floorplanning, placement, and routing, with each stage taking hours to weeks to complete. Discovering timing violations late in this flow forces costly iterations back to earlier stages, wasting computational resources and delaying product launches. While predicting post-routing timing from early-stage data could prevent these failures, existing machine learning approaches struggle with the massive abstraction gap between post-synthesis logical descriptions and post-routing physical layouts. We propose STEP-KD (Sequential Timing Evaluation via Progressive Knowledge Distillation), which leverages intermediate design stages as ``stepping stones'' for progressive knowledge transfer rather than attempting direct prediction. STEP-KD trains teacher models at the post-routing, post-placement, and post-floorplan stages, then sequentially distills their knowledge to a post-synthesis student model through representation alignment. Experiments on diverse circuits demonstrate that STEP-KD reduces timing prediction error compared to direct distillation and supervised baselines, and in most settings compared to the industry-standard Static Timing Analysis (STA) tool. STEP-KD reduces the weighted mean absolute percentage error of Total Negative Slack prediction to 19.78\%, compared with 74.84\% for STA. Our proposed method is step forward to identify timing problems earlier, avoiding expensive late-stage redesigns.
\end{abstract}


\section{Introduction}
The semiconductor industry underpins technological advances in AI/ML accelerators, autonomous vehicles, 5G/6G communications, and quantum computing. At the heart of this industry lies the integrated circuit (IC) design flow, which turns a register-transfer level (RTL) hardware description into a manufacturable layout through a sequence of stages: logic synthesis maps the RTL to a netlist of library cells, floorplanning and placement position these cells on the chip, and routing connects them with metal wires. Each stage can take hours to weeks for complex designs, and engineers tune many parameters along the way, such as clock frequency, cell sizes, placement density, and routing constraints. Together, these parameters determine whether the chip meets its timing requirements. Timing is verified with Static Timing Analysis (STA), the industry-standard method. STA can be run after every stage, but it only uses the information available at that stage: before routing, wire delays are unknown, cell delay models are simplified, and physical design constraints are not yet applied, so early-stage STA gives inaccurate estimates of the final, post-routing timing. The accurate timing is therefore known only after routing, and discovering violations at that point is expensive: engineers must iterate back to earlier stages, discarding days of computational work and delaying time-to-market by weeks or months.
 
The late discovery of timing problems presents an important machine learning challenge:
can we accurately predict late-stage timing quality from early-stage design data? We introduce Sequential Timing Evaluation via Progressive Knowledge Distillation (\textit{STEP-KD}) to address this challenge. STEP-KD is intended as an early prediction tool: right after logic synthesis, engineers obtain a fast estimate of the post-routing TNS and WNS and can flag designs likely to violate timing for parameter adjustments; relaxing the clock period, changing the placement density, or modifying the architecture, before committing to the expensive physical design flow. This enables design space exploration at synthesis speed rather than routing speed and avoids many late-stage iterations. Our key insight is that the intermediate design stages, specifically post-floorplan and post-placement, bridge the abstraction gap between our starting stage, post-synthesis, and our prediction target, post-routing. Intermediate stages provide progressively refined physical information that serves as ``stepping stones'' for knowledge transfer. Rather than attempting a single large leap, \textit{STEP-KD} transfers timing knowledge sequentially through adjacent stages. Each distillation step bridges a manageable gap with well-aligned feature spaces, enabling more effective knowledge transfer than direct approaches. This progressive distillation strategy mirrors the natural hierarchy of the IC design flow itself, where each stage incrementally adds physical details---placement provides cell coordinates, clock tree synthesis (CTS) adds clock network information, and routing completes the physical picture.
 
Our contributions are fourfold: (1)We propose the first progressive knowledge distillation approach for IC timing prediction that explicitly leverages intermediate design stages as sequential teachers, advancing beyond single-step distillation methods. (2) We demonstrate that this stage-sequential approach outperforms direct distillation baselines, validating our hypothesis that intermediate stages provide better representation alignment than attempting to bridge the full abstraction gap in one step. 
(3) We construct a dataset of about 12k circuit implementations of 922 RTL designs, each represented at four design stages together with the source RTL, which we will release publicly upon acceptance.  (4) We provide extensive experimental validation on diverse circuit designs, showing that STEP-KD enables accurate post-routing timing prediction from post-synthesis data. Taken jointly, our contributions offer a substantial step towards the goal of providing circuit engineers with instant feedback, allowing them to identify timing violations earlier in the design cycle and avoid costly late-stage iterations.
 
\section{Related Work}
 
\subsection{Circuit Timing Prediction}
Recent advances in machine learning for electronic design automation (EDA) have demonstrated strong potential for predicting design metrics and guiding decision-making across various stages of the physical design flow~\citep{graph-for-eda-1, graph-for-eda-2, hardware-logic-synthesis, ijcai2025p1033}. In particular, a growing body of work~\citep{circuitfusion, MasterRTL, RTLtimer-DAC24, llm-rtl-amik, net-tag, multimodal-timing} explores the use of graph-based and language-based models for predicting computationally intensive metrics such as power, performance, and area (PPA) early in the design flow. For PPA prediction at physical design stages, \citet{barboza2019machine} investigates machine learning techniques for post-routing net delay and slew prediction using net features extracted from placement results, but does not directly output global timing metrics. Building on this line of research, \citet{guo2022timing} presents a timing engine-inspired GNN framework that accurately predicts post-routing arrival times and slacks. However, both of these methods operate at the post-placement stage and are not readily extendable to earlier design stages.
 
Several recent studies have attempted to evaluate circuit quality at the early RTL stage. RTL-Timer~\citep{fang2024annotating} proposes a Boolean operator graph (BOG) representation of RTL designs and trains a model to predict post-synthesis endpoint arrival times. \citet{amik2025graph} achieves accurate TNS and WNS estimation at the RTL stage using an LLM-augmented GNN framework trained on synthesized RTL datasets. However, these approaches are confined to the logic synthesis stage and do not incorporate layout-aware timing features. To bridge the gap between layout and RTL, \citet{wang2025bridging} introduces a multi-granularity knowledge distillation framework that transfers physical information from a layout-aware teacher model to an RTL-level student model. NetTag~\citep{net-tag} proposes a layout-aware foundational netlist model through self-supervised pretraining and cross-stage alignment via contrastive learning. Despite these advances, by estimating timing at the RTL stage both approaches bypass the physical design process, in which diverse physical information emerges across the layout stages. They also predict timing before technology mapping, without the library-specific cell delays, drive strengths, pin capacitances and setup/hold timing arcs on which path delays fundamentally depend. We provide a detailed comparison of our work with previous related on in Appendix \ref{app:prior}.

\subsection{Privileged Knowledge Distillation in Machine Learning}

Our work also relates to the broader literature in machine learning on knowledge distillation between a teacher and a student model, when the teacher has access to more inputs than the student, known as privileged knowledge distillation.

An influential early approach on this topic for classification models is Generalized Distillation~\citep{LopezPaz15}, which proposes to train the student to match the gold labels, but adds a regularizer which encourages the student logits to match the teacher logits. Later work has also extended knowledge distillation beyond outputs. For example, \citet{TungM19} encourage internal activations of convolutional neural networks to match using a similarity loss regularizer, and \citet{aguilar2020knowledge} encourage attention weights and hidden vector representations of transformer networks to match using Kullback-Leibler and cosine similarity regularizers. Like these works, we also encourage internal representations to match between student and teacher models, although we employ graph neural networks.
Most related to ours is Teacher Assistant Knowledge Distillation~\citep{mirzadeh2020improved}, in which the authors introduce a ``teaching assistant'' that sits between the teacher and the student: the teacher teaches the TA, and the TA teaches the student. Our framework can be seen as an extension of this idea beyond three networks, although the focus of the authors was to bridge gaps in model capacity, whereas our method focuses on overcoming the information gaps between different stages.

\section{Background: Circuit Representation and Timing Analysis}\label{sec:background}
 
A synchronous digital circuit is represented as a directed graph $G = (V, E)$ whose nodes are circuit elements and whose edges are signal connections. The node set contains $(i)$ sequential elements (registers/flip-flops) or output gates $V_{\text{seq}} \subset V$, which serve as the start and end points of timing paths, and $(ii)$ combinational logic gates, which perform Boolean operations on their inputs. Each edge $(u, v)$ carries a delay $d(u,v)$ comprising the switching delay of the driving gate and the wire delay. The sequential elements are fixed across design stages, whereas the combinational logic may be altered from one stage to the next. STA decomposes the circuit into timing paths, each running from one sequential element through combinational logic to another, and computes the \textbf{slack} of every sequential element as the difference between the time its signal is required and the time it actually arrives, $\text{slack}(v) = \text{RAT}(v) - \text{AAT}(v)$ (see Appendix~\ref{app:timing-defs} for details). A positive slack is a timing margin; a negative slack is a \textbf{timing violation}, meaning the circuit fails at the target clock period. Because delays and clock timing are refined as the design progresses, slack estimated at an early stage can differ substantially from post-routing slack. We predict the post-routing slack of the sequential elements and summarize circuit timing with the \textbf{total negative slack} (TNS) and the \textbf{worst negative slack} (WNS),
\begin{align}
\label{eq:tns-wns}
\text{TNS} = \hspace{-5pt}\sum_{\substack{v \in V_{\text{seq}} \\ \text{slack}(v) < 0}}\hspace{-5pt}\text{slack}(v),
\qquad\text{WNS} = \min_{v \in V_{\text{seq}}} \text{slack}(v).
\end{align}
TNS quantifies the aggregate severity of all timing violations in the circuit, and WNS the single most critical one: a circuit can have a small WNS but a large TNS (many moderately failing paths), or a large WNS but a small TNS (one severely failing path).

\section{Method: STEP-KD}
\label{sec:method}
\begin{figure}[t]
  \centering
  \begin{subfigure}[b]{0.62\textwidth}
    \centering
    \includegraphics[width=\textwidth]{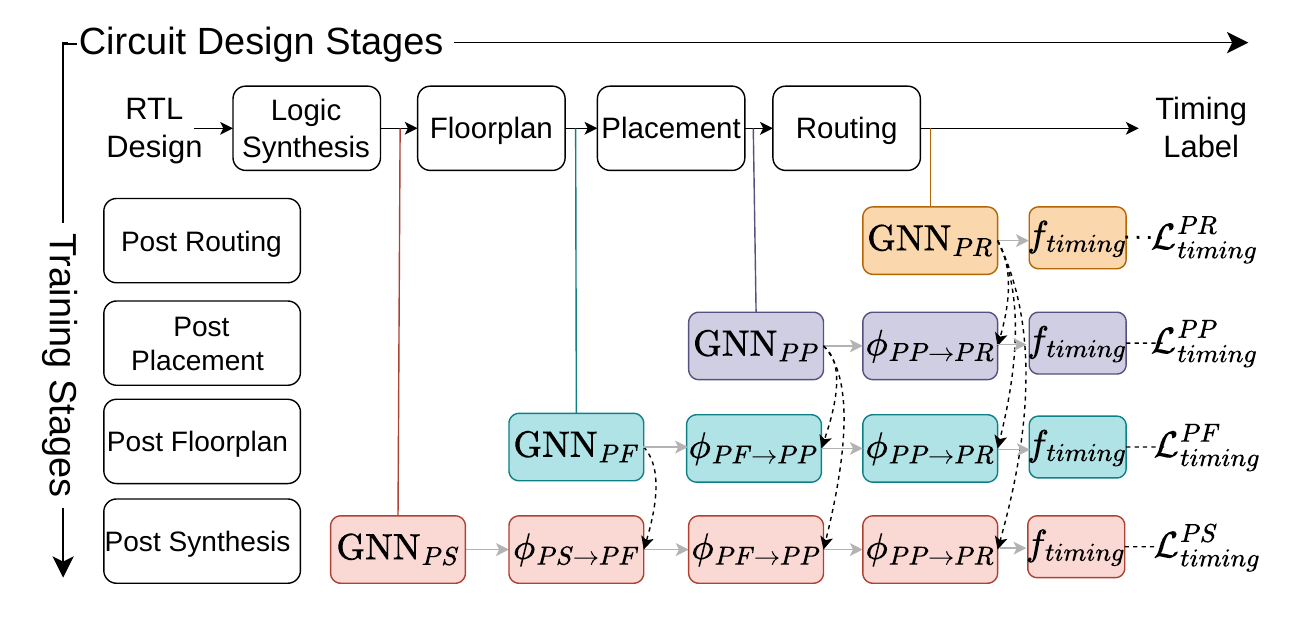}
    %\caption{}
    \label{fig:overview}
  \end{subfigure}
  \begin{subfigure}[b]{0.37\textwidth}
    \centering
    \includegraphics[trim={0, 0, 0, 0}, width=\textwidth]{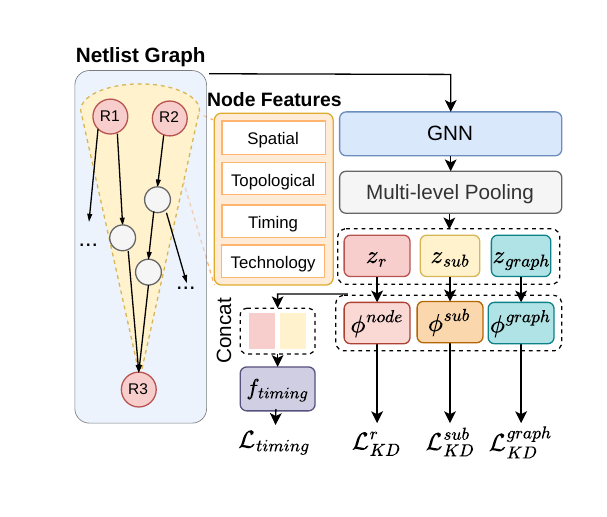}
    %\caption{}
    \label{fig:architecture}
  \end{subfigure}
  \caption{\textbf{Training flow (left)}: Models are trained one at a time down the design ladder, in the order PR, PP, PF, PS. Each model is trained to predict post-routing slack and, except for PR, is regularized by knowledge distillation (KD) losses that align its representations, passed through its own cascade of learned projections, with those of the frozen models at all later stages, which have access to physical information not yet available at its own stage. Only the post-synthesis model is used at inference. \textbf{Model architecture (right)}: The architecture of our machine learning model. It takes as input a graphical representation of a circuit at the post-synthesis stage (netlist graph) and outputs the slack for every sequential element in the circuit. From these predictions, circuit-wide metrics such as the total negative slack (TNS) and worst negative slack (WNS) can be computed.}
  \label{fig:combined_framework}
\end{figure}

We present STEP-KD, our method for predicting post-routing timing from post-synthesis circuit data. STEP-KD uses the same graph neural network architecture at every design stage (Section~\ref{sec:features}) and trains one model per stage, one at a time, from the latest stage to the earliest: each model is distilled from all later-stage models through a cascade of learned projections (Section~\ref{sec:training}). Once training is complete, only the post-synthesis model is kept for inference. Figure~\ref{fig:combined_framework} summarizes the training procedure and the model architecture.
 
\subsection{Features and Encoding}
\label{sec:features}
 
Our models take as input graph representations of circuits at four stages: post-synthesis (PS), post-floorplan (PF), post-placement (PP), and post-routing (PR). Each node carries a stage-dependent feature vector $\mathbf{x}_v \in \mathbb{R}^{d_{\text{in}}}$ that concatenates \textbf{spatial} features (physical coordinates $(x_v, y_v)$, available from post-placement onward and set to $(0,0)$ at the PS and PF stages), \textbf{timing} features (the current-stage slack from STA), \textbf{topological} features (one-hot cell type, depth-per-input, depth-per-output, and fan-in/fan-out counts), and \textbf{technology} features (drive strength, pin counts, library parameters, and the design-level placement density, core utilization and clock period); see Appendix~\ref{app:feature_details} for details. At each stage, a Graph Convolutional Network (GCN) with $L$ layers of neighborhood aggregation encodes these features into node embeddings $\mathbf{Z} = \{\mathbf{z}_v \in \mathbb{R}^{d_{\text{emb}}} : v \in V\}$, which capture both local neighborhood structure and global circuit properties.

\subsection{Multi-Granularity Representations and Projections}
\label{sec:representations}
\label{sec:multi_representation_pooling}
\label{sec:methodology-proj}
\label{sec:f_timing_description}

\textbf{Multi-granularity pooling.} From the node embeddings we extract representations at three granularities: the node-level embedding $\mathbf{z}_r$ of each sequential element $r \in V_{\text{seq}}$; a subgraph-level embedding $\mathbf{z}_r^{\text{sub}} = \frac{1}{|V_{\text{sub}}(r)|} \sum_{v \in V_{\text{sub}}(r)} \mathbf{z}_v$, obtained by mean pooling over the fan-in cone $G_{\text{sub}}(r) = (V_{\text{sub}}(r), E_{\text{sub}}(r))$ of $r$, i.e., all nodes and edges between $r$ and its source sequential elements; and a graph-level embedding $\mathbf{z}^{\text{graph}} = \frac{1}{|V|} \sum_{v \in V} \mathbf{z}_v$. At stage $s$ this gives $\mathbf{Z}^s = \{\mathbf{z}_r^s, \mathbf{z}_r^{s,\text{sub}}, \mathbf{z}^{s,\text{graph}} : r \in V_{\text{seq}}\}$, capturing individual sequential elements, their local combinational logic, and global circuit properties. The node-level representation also carries a learned
embedding of the current-stage STA slack (Appendix \ref{app:input_encoding}).

\textbf{Cascaded projections.} We index the stages in design order, $s_0 = \text{PS}$, $s_1 = \text{PF}$, $s_2 = \text{PP}$, $s_3 = \text{PR}$. A projection $\Phi_{s_k \to s_{k+1}}$ maps representations of one stage into the representation space of the next. It consists of three three-layer MLPs $\phi^{\text{node}}, \phi^{\text{sub}}, \phi^{\text{graph}}: \mathbb{R}^{d_{\text{emb}}} \to \mathbb{R}^{d_{\text{emb}}}$, one per granularity, each with a residual connection around every linear layer. The model at stage $s_k$ learns its own cascade of projections, one per design step, whose composition $\Phi_{s_k \to s_j} = \Phi_{s_{j-1} \to s_j} \circ \cdots \circ \Phi_{s_k \to s_{k+1}}$ maps its representations into the space of any later stage $s_j$. Each single projection $\Phi_{s_{k} \to s_{k+1}}$ thus bridges only one design step. For example, the PF model reaches the PR space through the PP space with the map $\Phi_{s_1 \to s_3} = \Phi_{s_{2} \to s_3} \circ \Phi_{s_1 \to s_{2}}$. For a (possibly composed) projection $\Phi$ with granularity components $\phi^{\text{node}}, \phi^{\text{sub}}, \phi^{\text{graph}}$, the alignment with target representations $\mathbf{Z}'$ is a weighted sum over granularities,
\begin{equation}
\begin{aligned}
\|\Phi(\mathbf{Z}) - \mathbf{Z}'\|^2 = \; & \frac{\beta_1}{m} \sum_{r \in V_{\text{seq}}} \|\phi^{\text{node}}(\mathbf{z}_r) - \mathbf{z}'_r\|_2^2 + \frac{\beta_2}{m} \sum_{r \in V_{\text{seq}}} \|\phi^{\text{sub}}(\mathbf{z}_r^{\text{sub}}) - \mathbf{z}'^{\text{sub}}_r\|_2^2 \\
& + \beta_3 \|\phi^{\text{graph}}(\mathbf{z}^{\text{graph}}) - \mathbf{z}'^{\text{graph}}\|_2^2,
\end{aligned}
\end{equation}
where $m = |V_{\text{seq}}|$ and $\beta_1, \beta_2, \beta_3$ are hyperparameters weighting the importance of each granularity. This multi-granularity alignment ensures that the student learns to match each teacher's representations at multiple levels of abstraction -- from individual sequential elements to local combinational logic to global circuit properties.

\textbf{Timing predictor.} The timing predictor $f_{\text{timing}}^{s}$ of the stage-$s$ model predicts the post-routing slack of each sequential element $r \in V_{\text{seq}}$ from its node- and subgraph-level representations after projection to the PR space. Writing $\tilde{\mathbf{z}}_{r}^{s} = \phi_{s \to \text{PR}}^{\text{node}}(\mathbf{z}_{r}^{s})$ and $\tilde{\mathbf{z}}_{r}^{s,\text{sub}} = \phi_{s \to \text{PR}}^{\text{sub}}(\mathbf{z}_{r}^{s,\text{sub}})$, where $\phi_{s \to \text{PR}}^{\text{node}}$ and $\phi_{s \to \text{PR}}^{\text{sub}}$ are the node- and subgraph-level components of $\Phi_{s \to \text{PR}}$ (set to identity for $s = \text{PR}$), the predictor computes
\begin{equation}
\widehat{\text{slack}}_{\text{PR}}(r) = f_{\text{timing}}^{s}\bigl( [ \tilde{\mathbf{z}}_{r}^{s};\; \tilde{\mathbf{z}}_{r}^{s,\text{sub}} ] \bigr),
\end{equation}
where $[\cdot;\cdot]$ denotes concatenation and $f_{\text{timing}}^{s}: \mathbb{R}^{2d_{\text{emb}}} \to \mathbb{R}$ is an MLP. The graph-level representation enters training only through the distillation loss. 

\subsection{Training}
\label{sec:training}

STEP-KD trains one model per design stage and climbs down the design ladder: PR first, then PP, PF and finally PS. All four models predict the same post-routing slack labels. Each model is trained only after every later-stage model has been trained and frozen, and it is distilled from all of them; the PR model, which has no later stage, is trained with supervision alone. The PP model thus learns from PR, the PF model from PP and PR, and the PS model from PF, PP and PR (Figure~\ref{fig:combined_framework}, left). Each teacher learns to predict post-routing timing metrics using data from its respective stage.

The model at stage $s_k$ predicts post-routing slack from its PR-space representations and aligns its projected representations with the encoder outputs of every frozen later-stage model,
\begin{equation}
\label{eq:loss}
\begin{aligned}
\mathcal{L}^{s_k} = \; & \frac{\gamma_{\text{timing}}}{|V_{\text{seq}}|} \sum_{r \in V_{\text{seq}}} \Big| \widehat{\text{slack}}_{\text{PR}}(r)- \text{slack}_{\text{PR}}(r) \Big| \\
& + \gamma_{\text{KD}} \sum_{j=k+1}^{3} \big\|\Phi_{s_k \to s_j}(\mathbf{Z}^{s_k}) - \mathbf{Z}^{s_j}\big\|^2.
\end{aligned}
\end{equation}
The KD sum is empty when training the PR model. During the PP model training, the KD term aligns the representations of the PP model with the trained PR model. During the PF model training, it aligns the representations of the model with the PP and PR models. Finally, during the PS model training, it aligns the student with the PF, PP and PR models after one, two and three projection steps. Because the PF and PP teachers were themselves distilled from later stages, their representations are already aligned with those of the stages above them, so knowledge from routing reaches the student one rung at a time. Table~\ref{tab:intermediate} compares this against independently trained teachers.

\subsection{Inference}
At inference, only the PS model is used: its representations pass through its projection cascade (PS $\to$ PF $\to$ PP $\to$ PR) to its timing predictor $f_{\text{timing}}^{\text{PS}}$. The timing metrics can then be calculated directly from the node-level timing predictions for each circuit using Equation~\eqref{eq:tns-wns}, with the predicted  $\widehat{\text{slack}}$ values from the post-synthesis graph. This gives an early estimate of the PR timing information, saving expensive iterations in the design flow. \footnote{The sequential nodes are in one-to-one correspondence across design stages, so the sums range over the same set of nodes.}

\section{Experiments}

\subsection{Dataset Generation}

To train and evaluate STEP-KD, we require circuit representations at all four design stages (post-synthesis, post-floorplan, post-placement, post-routing) for the same designs. We construct this dataset using open-source RTL designs and an automated EDA flow.

\textbf{Source RTL Designs.} We collect 24,200 Verilog designs from two sources: 18,400 examples from \citet{pei2024betterv} and 5,800 from \citet{Thakur2022BenchmarkingLL}. These designs originate from open-source GitHub repositories and hardware design textbooks, covering arithmetic units, control logic, memory controllers, and communication interfaces.

\textbf{Multi-Stage Circuit Generation.} We run each design through logic synthesis and the complete physical design flow using the open-source OpenROAD platform~\citep{ajayi2019openroad} with the ASAP7 7nm predictive PDK. To obtain diverse implementations, we sweep the placement density over $\{0.7, 0.75, 0.8\}$ and the core utilization over $\{40, 45, 50, 55, 60\}\%$, giving 15 configurations per design; the clock period is set to OpenROAD's default for each design and used as an input feature. Of the 363K attempted runs, approximately 6\% complete all four stages with valid timing analysis, the rest failing because of, e.g., routing congestion or placement density conflicts. This yields 22k circuit implementations. Using 72 Core Intel(R) Xeon(R) Gold 6140 CPUs, each of the 15 runs took about one week to complete. After removing circuits with fewer than 150 nodes at the post-synthesis stage as trivial, we obtain 922 designs and 12.2k circuit implementations, each represented at all four stages; statistics are given in Appendix~\ref{app:dataset}. We will release the full dataset, including the source RTL and the graph representations at all four stages, upon acceptance of this paper.

\textbf{Dataset Split and Labels.} We split the source circuits into a 70/10/20 train/validation/test split. The three sets consist of 8.6k, 1.2k and 2.2k circuits, respectively (with 650, 94, and 178 unique RTL designs). All circuits derived from the same RTL design belong to the same split, so models are evaluated on unseen designs rather than on parameter variations of seen ones. The labels at every stage are the post-routing slack values from OpenROAD's final static timing analysis report.

\textbf{Additional Benchmark.} We further evaluate all methods on circuits from the OpenLLM-RTL benchmark~\citep{10.1145/3676536.3697118} as a complementary test set. Specifically, we use its RTLLM 2.0 designs: 50 diverse, simple, hand-crafted RTL circuits spanning arithmetic, memory, control, and processor-oriented functionality, in contrast to our primary test set, which consists of larger and more structurally complex open-source RTL designs. This diversity in circuit functionality allows us to assess whether the proposed approach generalizes across different types of RTL designs. None of these circuits are used during training. We pass them through the same flow with the same parameters, obtaining 189 circuits of 38 unique designs.

\subsection{Training and Evaluation Setup}

\textbf{Model Architecture.} We implement our graph neural network encoder using PyTorch Geometric. The GCN consists of 3 message-passing layers with a hidden dimension of 128, ReLU activation functions, and layer normalization. For the projection networks $\Phi_{s_1 \to s_2}$, each three-layer MLP uses a hidden dimension of 128, ReLU activations, and a residual connection around each linear layer, which we find lowers the alignment loss and improves STEP-KD performance. The timing predictor (Section~\ref{sec:f_timing_description}) is implemented as a 2-layer MLP.

\textbf{Training Configuration.} We train all models using the Adam optimizer with an initial learning rate of $10^{-4}$. We employ a ReduceLROnPlateau scheduler with patience of 20 epochs and a decay factor of 0.5. We train all models until convergence and keep the checkpoint with the lowest validation loss. For STEP-KD, the models are trained one at a time in the order PR, PP, PF, PS (Section~\ref{sec:training}), and each model is frozen before the next one is trained.

\textbf{Evaluation Protocol.} Each of the 178 test designs has several implementations (different placement density and core utilization). We conduct 20 evaluation runs, each sampling one implementation per design, and report the mean and standard deviation over these runs of the median absolute error (MedAE, in the units of the slack labels), WMAPE (Weighted Mean Absolute Percentage Error), and RSE (Relative Squared Error); Table~\ref{tab:main-r2} in the appendix additionally reports $R^2$ and the mean absolute error (MAE). The three main metrics differ deliberately in their sensitivity to outliers. MedAE ignores them: an anomaly does not shift the median unless it affects more than half of the circuits. WMAPE divides the total absolute error by the total magnitude of the labels, so it is robust to large relative errors on circuits with small slack but sensitive to large errors on circuits with large slack. RSE squares the errors, so a single large mistake can dominate it.

\subsection{Baselines and Model Variants}

We compare STEP-KD against three baselines and one ablated variant.

\textbf{Post-Synthesis STA.} OpenROAD's static timing analysis at the post-synthesis stage, the standard industry practice for early-stage timing estimation.

\textbf{No KD.} A GNN with the same encoder, pooling and timing predictor as STEP-KD, trained on post-synthesis graphs with the timing loss alone. It tests whether direct supervised prediction can bridge the synthesis-to-routing gap without intermediate-stage information.

\textbf{Direct KD.} Single-step distillation from one frozen teacher at the PR, PP or PF stage to the post-synthesis student through a single multi-granularity projection, similar to RTLDistil~\citep{wang2025bridging}. Each teacher is trained independently without a KD loss. With the PR teacher it attempts to bridge the full synthesis-to-routing gap in one step, isolating the benefit of progressive over direct distillation.

\textbf{STEP-KD w/o sequential KD.} Identical to STEP-KD for the PS student, but its PF and PP teachers are trained independently rather than distilled from later stages, isolating the contribution of distilling the teachers themselves.

\section{Results}
 
\input{tables/main_combined.tex}

\begin{figure}[t]
\centering
\setlength{\tabcolsep}{1pt}
\begin{tabular}{ccccc}
 & \small STA, TNS & \small STEP-KD, TNS & \small STA, WNS & \small STEP-KD, WNS \\
\rowlab{Held-out} & \scat{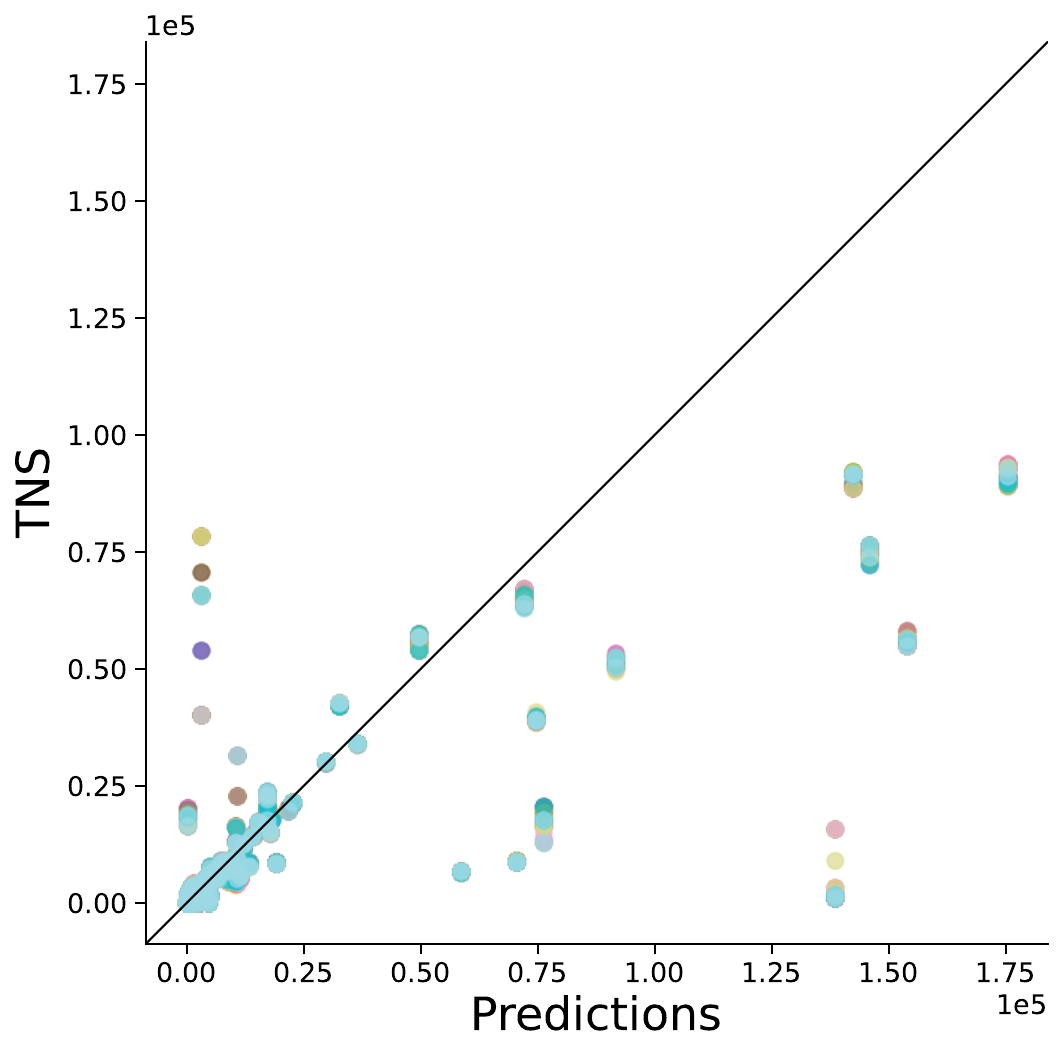} & \scat{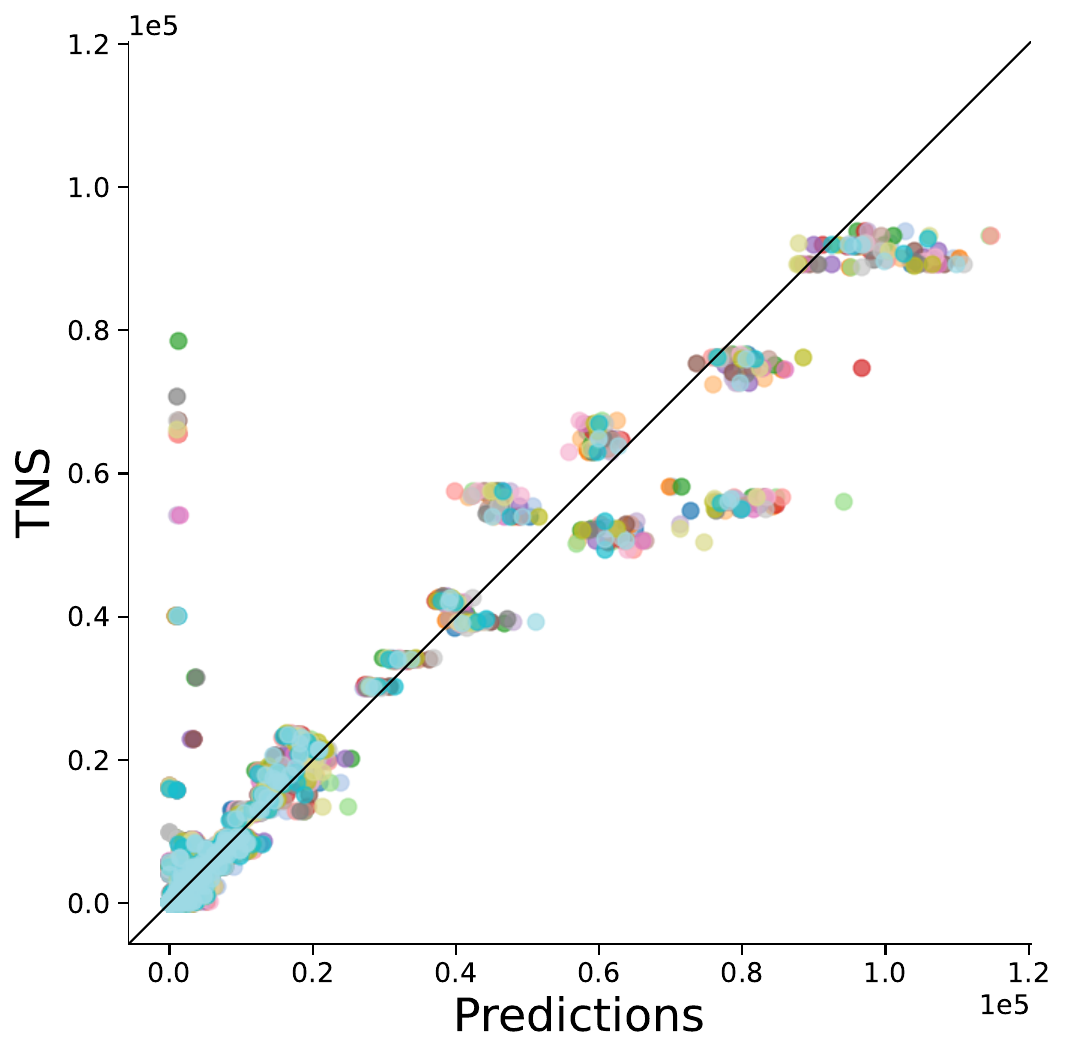} & \scat{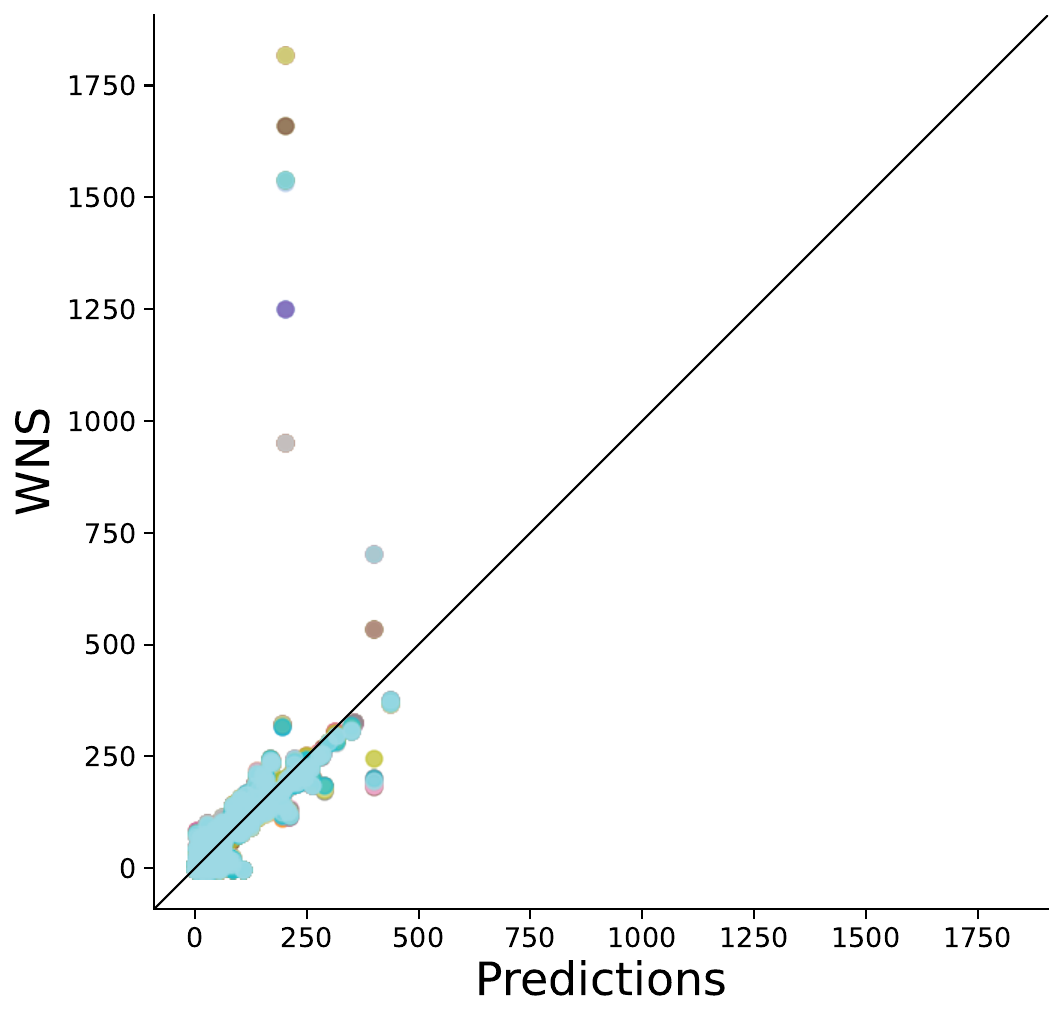} & \scat{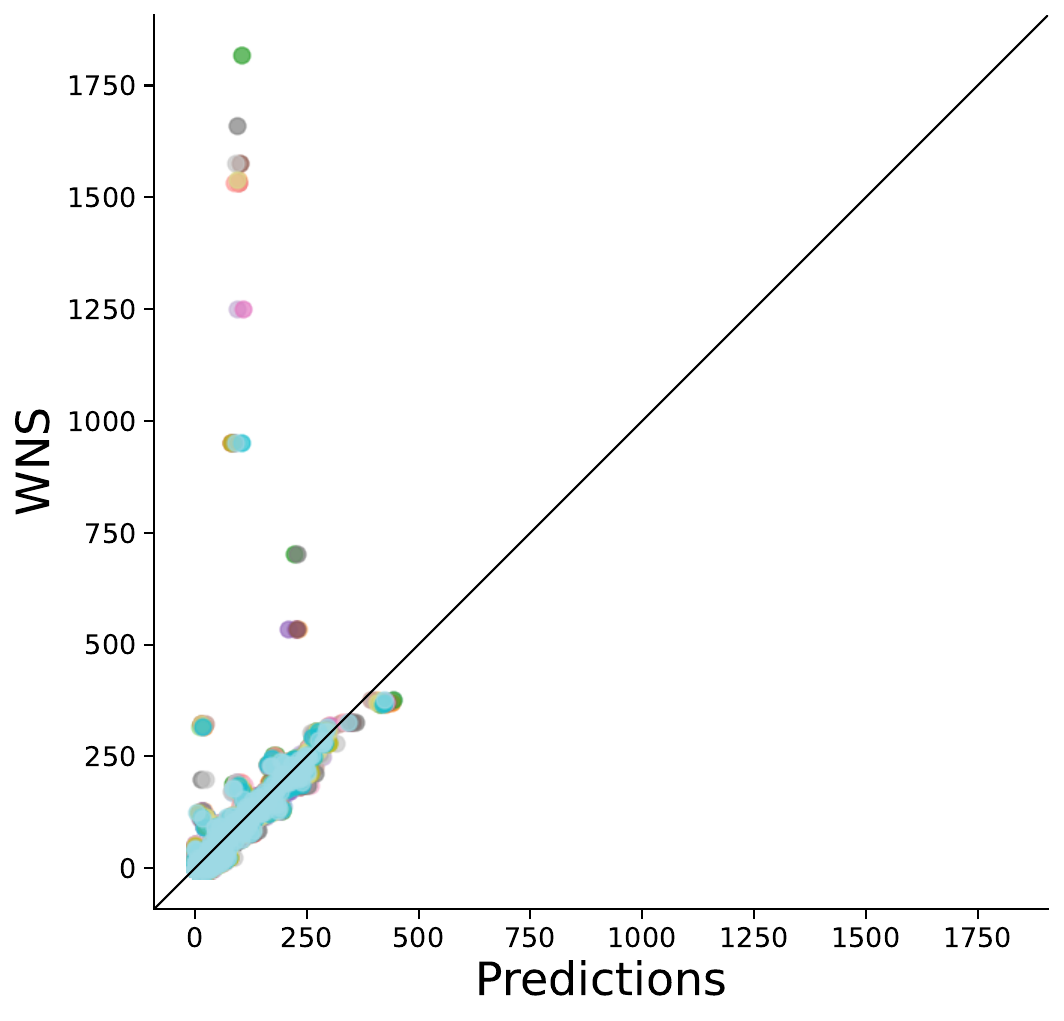} \\
\rowlab{OpenLLM-RTL} & \scat{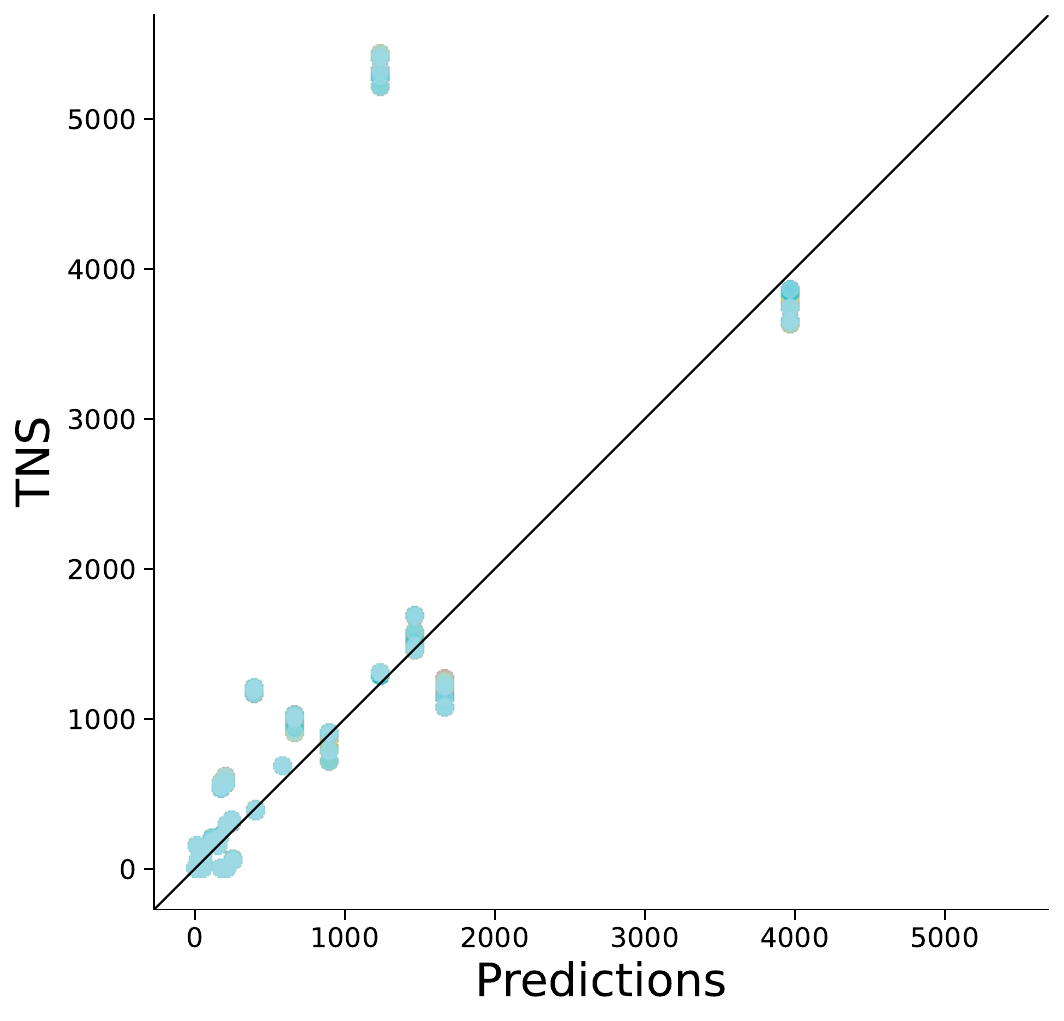} & \scat{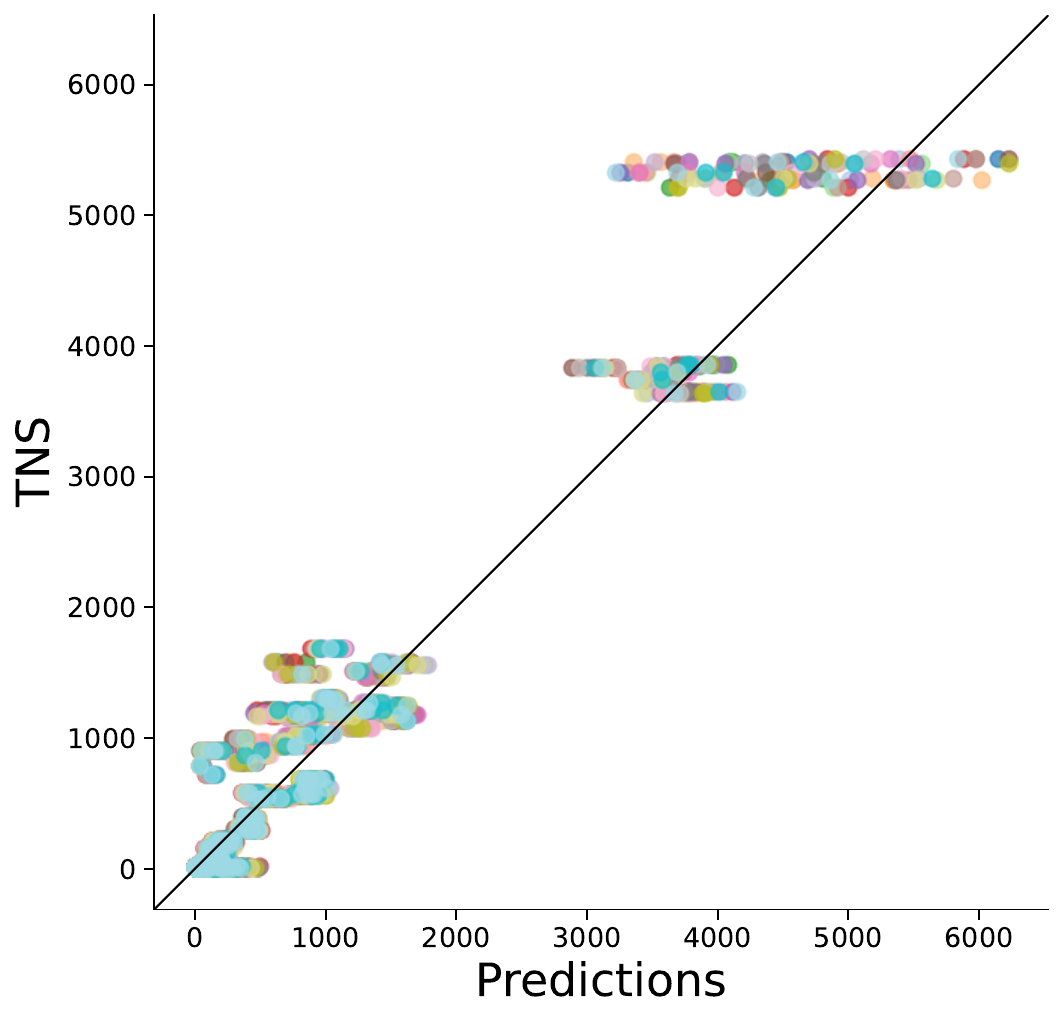} & \scat{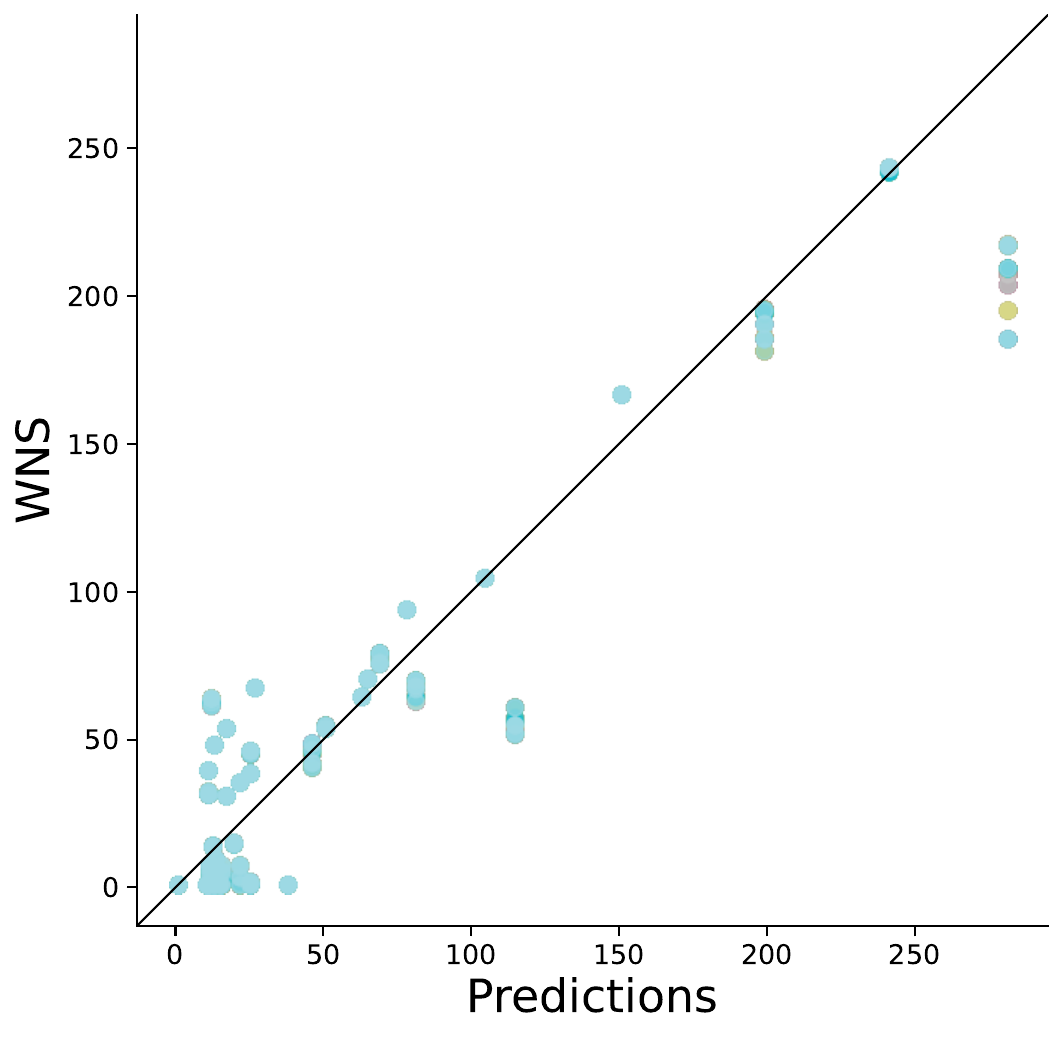} & \scat{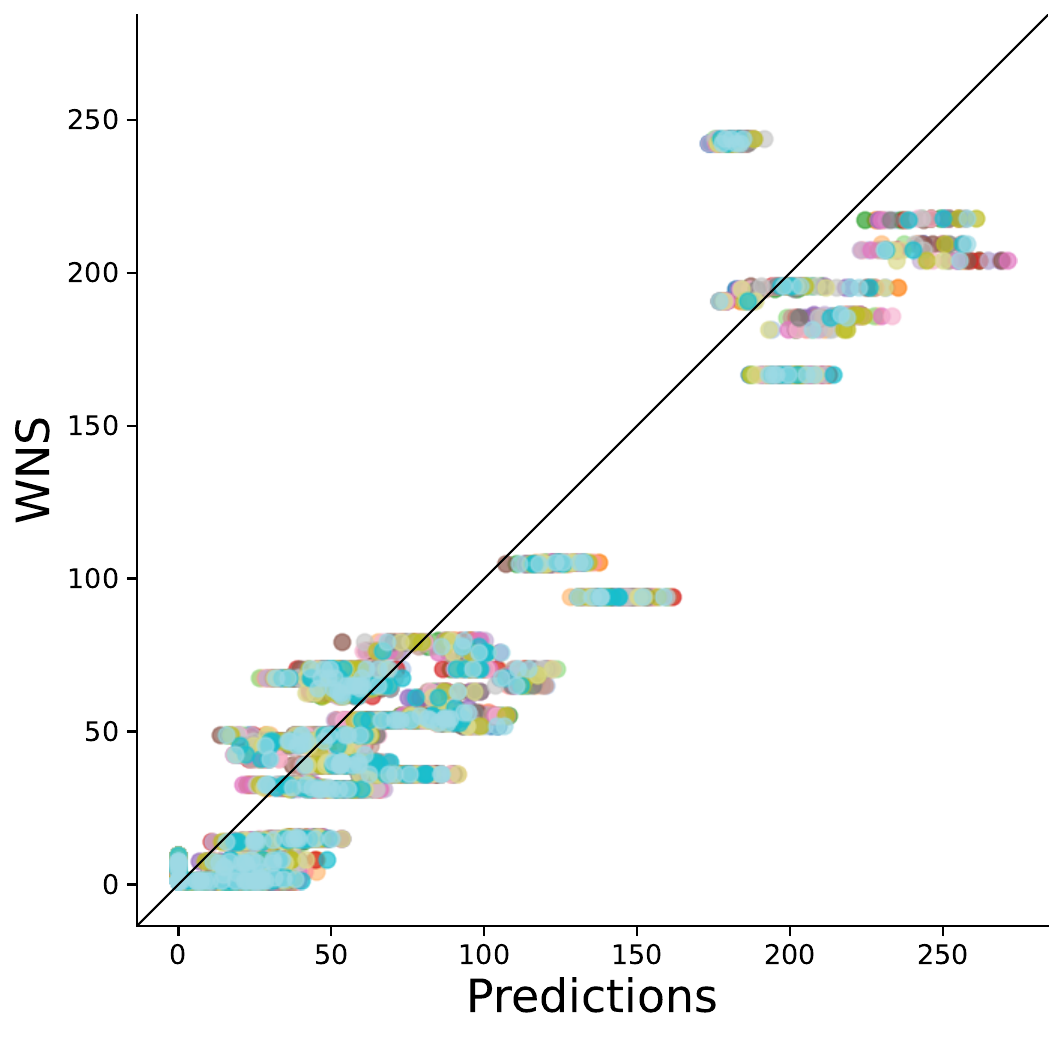} \\
\end{tabular}
\caption{Predicted versus post-routing TNS and WNS (magnitudes) for post-synthesis STA and STEP-KD on the held-out test set (top) and the OpenLLM-RTL benchmark (bottom). The diagonal marks perfect prediction. Each color is one of the 20 evaluation runs, each of which samples one implementation per design.}
\label{fig:scatter}
\end{figure}
 
\textbf{Main results.} Table~\ref{tab:main} summarizes the comparison. On the held-out test set, STEP-KD achieves the lowest TNS and WNS error on all three metrics. On both datasets it outperforms No KD and every Direct KD variant on every metric, showing that distilling through the intermediate stages is more effective than distilling from any single stage. On OpenLLM-RTL, STEP-KD has the lowest TNS WMAPE and RSE, and on WNS it is on par with post-synthesis STA, which is considerably more accurate on this benchmark than on the held-out set. Without distillation, the learned model is less accurate than STA on WNS on both datasets, so distillation is essential for this metric.
 
\textbf{Robustness to outliers.} Comparing the three metrics, which differ in their sensitivity to outliers (Section~5.2), together with the scatter plots in Figure~\ref{fig:scatter} shows how the errors are distributed. On the held-out set, the WNS RSE of every method is high and varies strongly across evaluation runs, even though STEP-KD's WNS MedAE and WMAPE are low. The scatter plots explain this: a handful of circuits have a post-routing WNS far larger than predicted from the post-synthesis netlist (upper left of the WNS panels), and neither STA nor STEP-KD anticipates this jump; similar outliers appear for TNS. WNS is intrinsically harder to predict than TNS: as the minimum slack over all sequential elements, it is set by a single critical path, so any error on that one path carries over entirely to WNS, whereas TNS aggregates the slack of many endpoints. The circuits of OpenLLM-RTL are much simpler, so their timing changes little between post-synthesis and post-routing, and post-synthesis STA is a reasonable estimate: it has the lowest TNS MedAE and is on par with STEP-KD on WNS. Nonetheless, STEP-KD has a far lower TNS WMAPE and RSE, because STA grossly underestimates a few circuits with large TNS (e.g., near 5{,}000) that STEP-KD predicts much more closely. Appendix~\ref{app:stage_jumps} also shows that STEP-KD, unlike STA, adapts its predictions to the physical design parameters of each implementation.

\input{tables/heldout_intermediate.tex}

\textbf{Teacher models and sequential distillation.} Table~\ref{tab:intermediate} reports the models at the intermediate stages. On the held-out set, the teachers are more accurate than STA at their own stage on every metric, and distilling them along the ladder further improves their TNS predictions, with a mixed effect on WNS. The effect is largest at the student: aligning the PS model to ladder-distilled teachers instead of independently trained ones improves all six metrics. Without sequential distillation, the PS student has a higher TNS WMAPE and RSE than every Direct KD variant, so the benefit of the cascade largely comes from passing knowledge down the ladder rather than from aligning to more teachers. %On OpenLLM-RTL (Table~\ref{tab:intermediate-olr}), sequential distillation improves the PF teacher and the student's TNS but degrades the PP teacher, and it lowers the student's WNS RSE at the cost of a slightly higher WNS MedAE and WMAPE.
 
\begin{figure}[t]
\centering
\begin{minipage}[t]{0.49\textwidth}
  \centering
  \includegraphics[width=\textwidth]{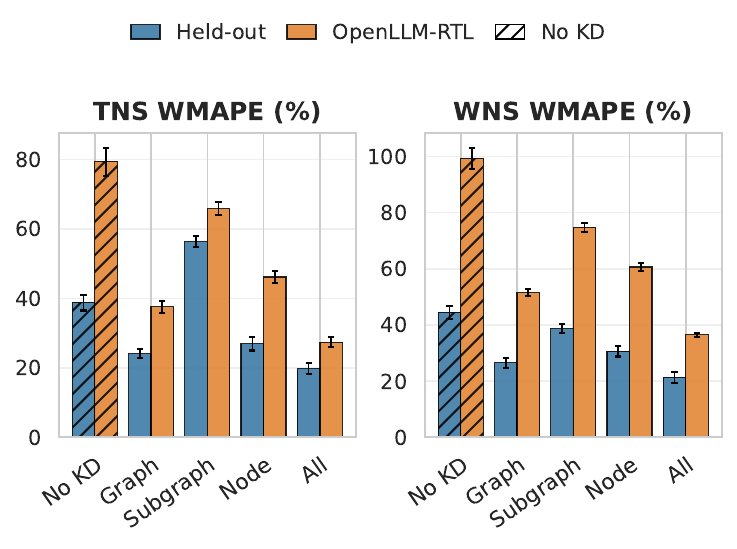}
  \caption{Impact of multi-granularity distillation on WMAPE (lower is better). Each single-granularity variant distills from the same teachers as STEP-KD (\emph{All}) but aligns only the listed granularity; hatched bars show No KD.}
  \label{fig:multigranularity_ablation}
\end{minipage}\hfill
\begin{minipage}[t]{0.49\textwidth}
  \centering
  \includegraphics[width=\textwidth]{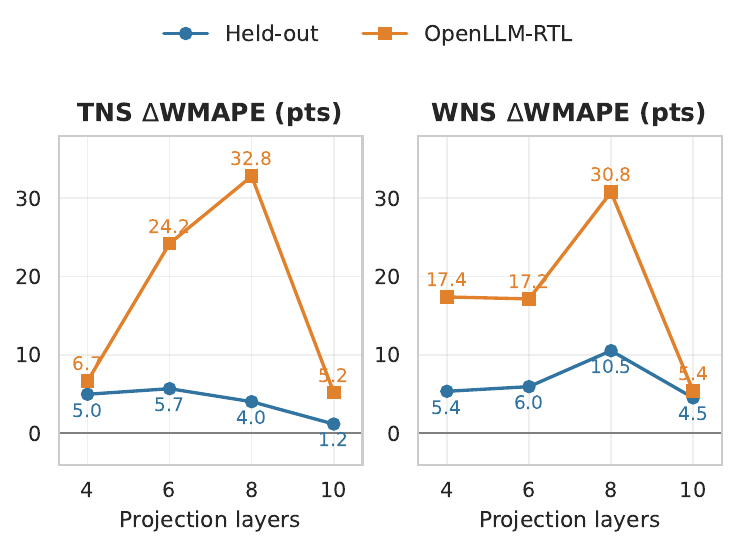}
  \caption{WMAPE of Direct KD (PR teacher) with 4 to 10 projection layers minus the WMAPE of STEP-KD, in percentage points. Positive values mean STEP-KD has lower error.}
  \label{fig:depth}
\end{minipage}
\end{figure}
 
\textbf{Impact of multi-granular distillation.} To measure the contribution of each granularity, we train PS students that distill from the same teachers as STEP-KD but align only the graph-, subgraph- or node-level representations. Aligning all three granularities gives the lowest error on both datasets (Figure~\ref{fig:multigranularity_ablation}), and on every metric (Tables~\ref{tab:app-granularity} and~\ref{tab:app-granularity-olr}). Graph-level alignment is the most effective single granularity, followed by node-level alignment. Subgraph-level alignment alone is the weakest and on held-out TNS is even worse than No KD, yet it helps in combination with the other two, so the granularities are complementary.
 
\textbf{Effect of projection network depth.} STEP-KD's PS model composes three 3-layer projections. To test whether its advantage comes from this added capacity rather than from the design ladder, we vary the depth of the single projection of Direct KD (PR teacher) from 4 to 10 layers. STEP-KD is more accurate than Direct KD at every depth, on both datasets and every metric (Figure~\ref{fig:depth}; Tables~\ref{tab:app-depth} and~\ref{tab:app-depth-olr}). Deeper projections do not reliably narrow the gap and can be unstable, and even the 10-layer projection, deeper than STEP-KD's 9-layer cascade, still trails it. The advantage of STEP-KD therefore does not come from depth alone.
 
%\textbf{Timing predictor inputs.} We ablate the inputs of the timing predictor, including a slack-bias variant that additionally feeds the current-stage STA slack to the predictor, in Appendix~\ref{app:slack_bias}.
 
\section{Conclusion}
We presented STEP-KD, a novel approach for predicting post-routing timing from post-synthesis circuit data. Our key insight is that leveraging intermediate design stages, post-floorplan and post-placement, as sequential teachers enables more effective knowledge transfer than direct distillation. STEP-KD can potentially enable engineers to identify timing violations earlier in the design cycle, helping to cut costs and avoiding expensive late-stage redesigns.

 \iffalse
% ------------------------------------------------------------------
% Statements (do not count toward the page limit)
% ------------------------------------------------------------------
\subsection*{AI use statement}
% Required by ICLR 2027. See the ICLR 2027 AI Policy for Authors.
All code implementing the methods and constructing the dataset was written by the authors, without AI-generated code. The main paper body was written by the authors; Large Language Models were used to polish and revise the text and to help produce figures and tables from our experimental results. The authors reviewed all AI-assisted text, figures and tables and take full responsibility for the content of the paper.
 
\subsection*{Reproducibility statement}

Section~5.1 describes the construction of the dataset, Section~5.2 the model architecture, training configuration and evaluation protocol, and Appendix~\ref{app:feature_details} the input features. The full dataset, including the source RTL and the graph representations at all four design stages, will be released upon acceptance.

\fi

\clearpage
% Numerical natbib style shipped with the arXiv preprint template.
% For an author--year variant, switch this to plainnat.
\bibliographystyle{unsrtnat}
\bibliography{references}

\clearpage
\appendix

\section{Circuit Timing Definitions}\label{app:timing-defs}
This appendix details the timing concepts summarized in Section~\ref{sec:background}.
 
\textbf{Timing paths.} A timing path $P = (v_0, v_1, \ldots, v_k)$ is a sequence of nodes connected by directed edges, starting at a register or primary input $v_0 \in V_{\text{seq}}$ and ending at a register or primary output $v_k \in V_{\text{seq}}$. The intermediate nodes $v_1, \ldots, v_{k-1}$ represent combinational logic. The \textbf{path delay} is the cumulative delay along the path,
\begin{equation}
\text{delay}(P) = \sum_{i=0}^{k-1} d(v_i, v_{i+1}),
\end{equation}
where each edge delay $d(u,v)$ includes the gate delay (the intrinsic switching time of the driving gate) and the interconnect delay (wire resistance-capacitance effects). In a synchronous circuit operating with clock period $T_{\text{clk}}$, signals must propagate through combinational logic and arrive at the destination register before the next clock edge; this is the fundamental timing requirement the circuit must satisfy. STA decomposes a design into timing paths and checks whether each signal meets this requirement, using the delays available at the current design stage. The accuracy of the resulting arrival times, and hence of the slack, depends on the accuracy of the delay values and clock timing information, which improve progressively through the design flow.
 
\textbf{Slack.} The slack at a node $v$, $\text{slack}(v) = \text{RAT}(v) - \text{AAT}(v)$, is the difference between its required arrival time (RAT) and actual arrival time (AAT), defined below. A positive slack indicates that the signal arrives earlier than required: the path has a timing margin and could tolerate additional delay. A negative slack means the signal arrives later than required, a timing violation: the circuit will fail to operate correctly at the target clock frequency.
 
\textbf{Actual Arrival Time (AAT)}
 
The \textbf{actual arrival time} (AAT) at a node $v$, denoted $\text{AAT}(v)$, represents the time at which a signal actually arrives at that node relative to the clock edge that launches it. For a sequential element node, AAT is initialized to the clock arrival time plus the clock-to-q delay (the time for the sequential element output to become valid after the clock edge):
\begin{equation}
\text{AAT}(v) = t_{\text{clk}}(v) + t_{\text{cq}}(v), \quad \text{for } v \in V_{\text{seq}}
\end{equation}
 
For a combinational node $v \in V_{\text{comb}}$, the AAT is computed by considering all possible input paths and taking the maximum (latest) arrival:
\begin{equation}
\text{AAT}(v) = \max_{u \in \text{fanin}(v)} \{\text{AAT}(u) + d(u,v)\}
\end{equation}
where $\text{fanin}(v) = \{u \in V : (u,v) \in E\}$ is the set of nodes that drive $v$. This max operation reflects the critical constraint that a combinational gate cannot produce a valid output until all its inputs have arrived and stabilized. The AAT propagates forward through the circuit graph from sequential elements through combinational logic to downstream sequential elements.
 
\textbf{Required Arrival Time (RAT)}
 
The \textbf{required arrival time} (RAT) at a node $v$, denoted $\text{RAT}(v)$, represents the latest time at which a signal must arrive at that node to meet timing constraints. This is computed through backward propagation from sequential elements. For a destination sequential element node $v \in V_{\text{seq}}$, the RAT is determined by the clock period and setup time requirement:
\begin{equation}
\text{RAT}(v) = t_{\text{clk}}(v) + T_{\text{clk}} - t_{\text{setup}}(v)
\end{equation}
where $t_{\text{clk}}(v)$ is the clock arrival time at register $v$, $T_{\text{clk}}$ is the clock period, and $t_{\text{setup}}(v)$ is the setup time (the minimum time before the clock edge that data must be stable). For a combinational node $v \in V_{\text{comb}}$, the RAT is computed by considering all fanout paths and taking the minimum (earliest) required time:
\begin{equation}
\text{RAT}(v) = \min_{w \in \text{fanout}(v)} \{\text{RAT}(w) - d(v,w)\}
\end{equation}
where $\text{fanout}(v) = \{w \in V : (v,w) \in E\}$ is the set of nodes driven by $v$. The RAT propagates backward from registers through combinational logic to upstream registers.

\section{Dataset Details}\label{app:dataset}
\subsection{Dataset Statistics}
\begin{figure}[b!]
\centering
\includegraphics[width=0.45\textwidth]{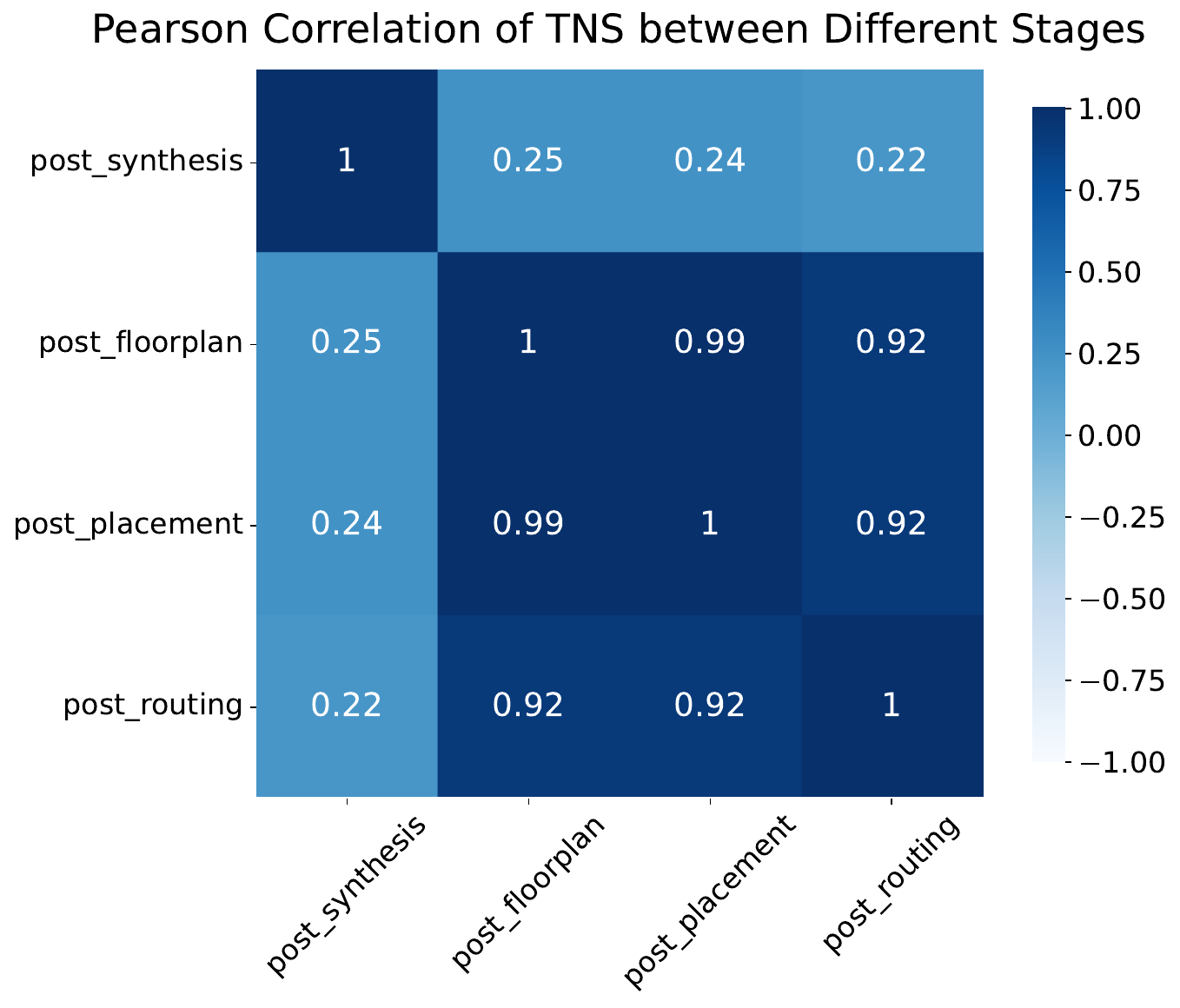}
\includegraphics[width=0.45\textwidth]{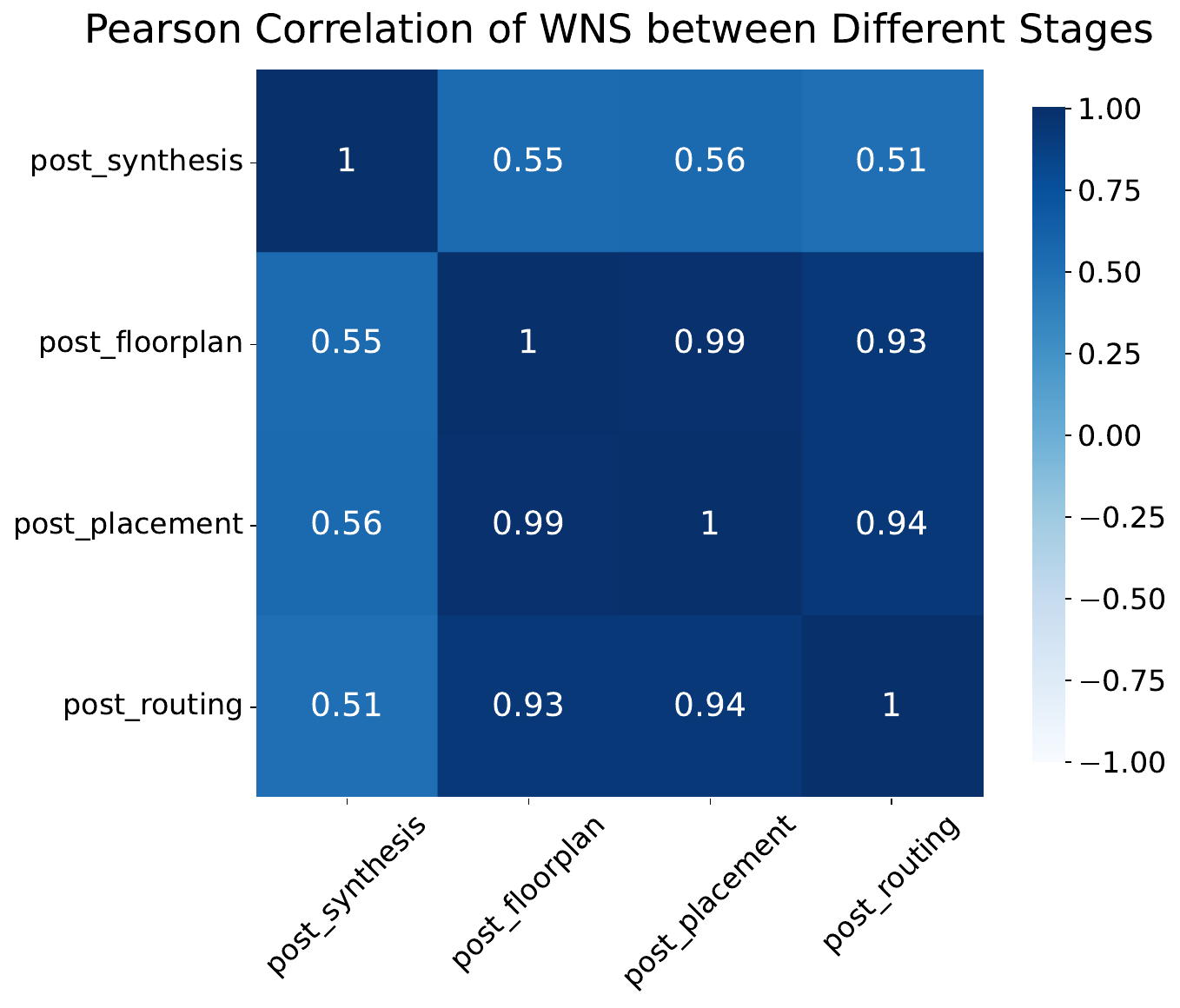}
\caption{Label Correlation Across Design Stages}
\label{fig:corr tables}
\end{figure}
 
\begin{figure}[t!]
\centering
 \includegraphics[width=0.32\textwidth]{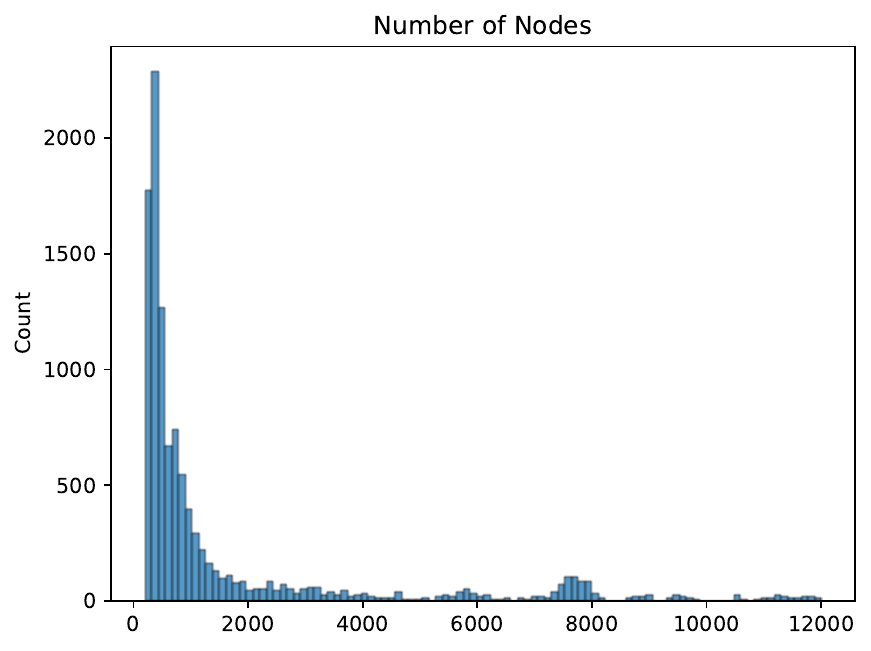}
 \includegraphics[width=0.32\textwidth]{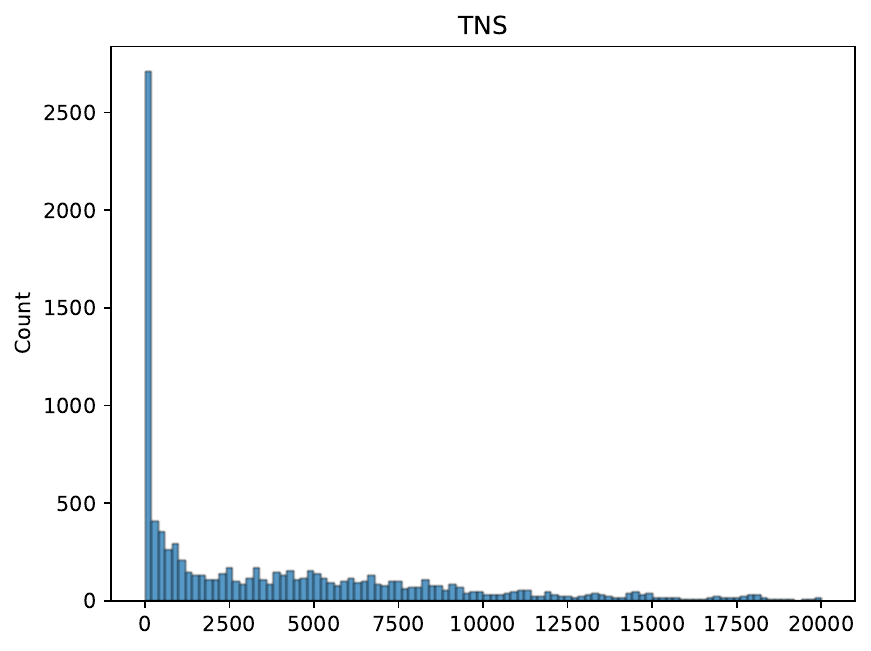}
 \includegraphics[width=0.32\textwidth]{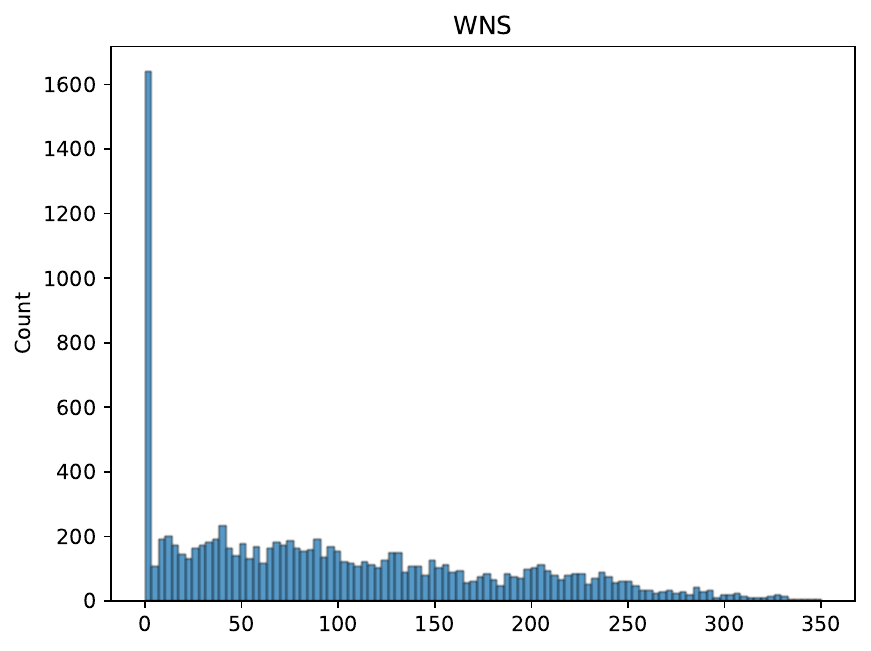}
  \caption{Histogram of Dataset Attributes}
  \label{fig:histograms}
\end{figure}

We plot the histogram of label and graph size distribution across the dataset in Figure~\ref{fig:histograms}. The TNS values fall between 0 and 538124, and the WNS values fall in the range of 0 to 2155. About 12 percent of circuits in the dataset have WNS/TNS values equal to zero.
 
Figure~\ref{fig:corr tables} depicts the correlation between timing labels from different stages. The largest change in labels occurs after floorplanning, while placement does not change the STA produced labels significantly.

\subsection{OpenLLM-RTL Benchmark}
From the OpenLLM-RTL benchmark, we exclude five circuits whose post-routing TNS and WNS are very large and identical to their post-synthesis values, so that post-synthesis STA is exact on them. Because of their magnitude, these circuits would dominate the WMAPE and RSE of every method and drive all results to near-perfect scores, masking the differences between methods.
 
\subsection{Data Feature Details}
\label{app:feature_details}
\begin{itemize}
    \item \textbf{Spatial Features: } These are the node coordinates, which become available after placement. We extract the coordinates for the post-placement and post-routing circuits, and normalize them per circuit by the maximum (x,y) coordinate in the circuit (so the values fall between zero and one). We assign coordinates (0,0) to all nodes in the post-synthesis and post-floorplan stages.
    \item \textbf{Timing Features: } The timing feature is a single scalar, which specifies the slack value estimated by the current stage STA.
    \item \textbf{Topological Features: } We extract 61 topological and functional features from the technology library reports. These features include \textit{area, fan in, fan out, SI direction, SN direction, capacitance, etc.}
\end{itemize}

\section{Input Encoding and Slack Bias}
\label{app:input_encoding}
 
\textbf{Input encoding.} Each node has four feature vectors: the spatial, timing and topological features described in Appendix~\ref{app:feature_details}, and the technology features (Section~\ref{sec:features}). Each vector is passed through a separate linear layer, and the four outputs are concatenated to form the input feature $\mathbf{x}_v \in \mathbb{R}^{d_{\text{in}}}$ of the node.
 
\textbf{Slack bias.} The GNN encoder outputs a representation $\mathbf{z}_v$ for every node. Before the pooled node-, subgraph- and graph-level representations are passed to the projection cascade and the timing predictor, we add to the node-level representation of each sequential element a learned embedding of its current-stage STA slack,
\begin{equation}
\mathbf{z}_r \leftarrow \mathbf{z}_r + p\big(\text{slack}(r)\big),
\end{equation}
where $p: \mathbb{R} \to \mathbb{R}^{d_{\text{emb}}}$ is a learnable projector. The term $p(\text{slack}(r))$ acts as a latent bias that passes the slack information directly to the timing predictor. We found that it improves both accuracy and training stability.
 
\input{tables/slack_bias.tex}

\textbf{Ablation.} Table~\ref{tab:slack_bias} compares three choices of the node-level representation for the post-synthesis model trained with Direct KD from the PR teacher: (i) the GNN output $\mathbf{z}_r$ alone, (ii) the slack embedding $p(\text{slack}(r))$ alone, and (iii) their sum, which we use throughout the paper. In variant (ii), the subgraph- and graph-level representations are unchanged and therefore still use all input features. Removing the slack embedding (i) degrades every metric sharply: TNS WMAPE rises from 24.05\% to 78.78\%, and WNS RSE exceeds 100\%, i.e., the predictions are less accurate than the mean. The current-stage slack is thus the most informative input for predicting post-routing slack. Although it is also one of the GNN's input features, message passing mixes it with the features of neighboring nodes, which is a likely reason the model cannot recover it precisely from $\mathbf{z}_r$; the slack embedding passes it to the timing predictor directly. Conversely, using the slack embedding without the GNN output (ii) comes close to the full model. The sum (iii) is better on five of the six metrics, most clearly on TNS MedAE (399.07 versus 495.26), although the other differences are within one standard deviation. The GNN node representation therefore adds a smaller refinement on top of the slack, mainly for the typical circuit.

 \section{Comparison with Prior Timing Prediction Methods}
\label{app:prior}
 
Table~\ref{tab:prior} summarizes how STEP-KD differs from the prior timing prediction methods discussed in Section~2.1.
 
\begin{table}[h]
\centering
\caption{Differences between STEP-KD and prior timing prediction methods.}
\label{tab:prior}
\resizebox{\textwidth}{!}{%
\begin{tabular}{llll}
\toprule
Method & Input stage & Prediction target & Training paradigm \\
\midrule
RTL-Timer~\citep{fang2024annotating} & RTL (pre-synthesis) & Post-synthesis timing & Supervised \\
NetTAG~\citep{net-tag} & Post-synthesis netlist & Slack and other tasks & Self-supervised pre-training \\
CircuitFusion~\citep{circuitfusion} & RTL (multimodal) & TNS/WNS and other tasks & Self-supervised pre-training \\
STEP-KD (ours) & Post-synthesis netlist & Post-routing TNS/WNS & Supervised sequential KD \\
\bottomrule
\end{tabular}}
\end{table}

 Besides the fact that our core contribution is different from prior work (such as RTL-Timer, NetTAG or CircuitFusion), these prior work also differ from STEP-KD in input stage, prediction target and training paradigm: RTL-Timer and CircuitFusion operate on RTL rather than on the post-synthesis netlist, RTL-Timer predicts post-synthesis rather than post-routing timing, and NetTAG and CircuitFusion rely on self-supervised pre-training and requires pre-training on a large corpus before fine-tuning on downstream tasks. % NetTAG is the closest to STEP-KD, as it also operates on the post-synthesis netlist, but it differs in two respects. First, it uses a multimodal representation: netlists are formulated as text-attributed graphs in which each gate is annotated with its symbolic logic expression and physical characteristics, and these are encoded by an LLM together with a graph transformer. Second, it requires pre-training on a large corpus before fine-tuning on downstream tasks, whereas STEP-KD is trained end-to-end on a single dataset.

% A comparison would conflate these factors with the contribution we study, so our baselines instead share STEP-KD's input, target, encoder and training data, and differ only in how they use information from the later design stages.

%NetTAG is the closest to STEP-KD, as it also operates on the post-synthesis netlist, but it differs in three respects. First, it uses a multimodal representation: netlists are formulated as text-attributed graphs in which each gate is annotated with its symbolic logic expression and physical characteristics, and these are encoded by an LLM together with a graph transformer. Second, it is a self-supervised foundation model trained with four pre-training objectives, whereas STEP-KD is trained with supervised sequential distillation, so a comparison would conflate the training paradigm with the architectural contribution. Third, it requires pre-training on a large corpus before fine-tuning on downstream tasks, whereas STEP-KD is trained end-to-end on a single dataset; a fair comparison would require pre-training NetTAG on our data under identical conditions, a substantial implementation effort. We therefore isolate the effect of progressive distillation with baselines that share STEP-KD's input, prediction target, encoder and training data (Section~5.3).

\section{Complete Scatter Plots}
\label{app:stage_jumps}

\begin{figure}[t]
\centering
\setlength{\tabcolsep}{1pt}
\begin{tabular}{ccccc}
 & \small PF STA, TNS & \small PP STA, TNS & \small PF STA, WNS & \small PP STA, WNS \\
\rowlab{Held-out} & \scat{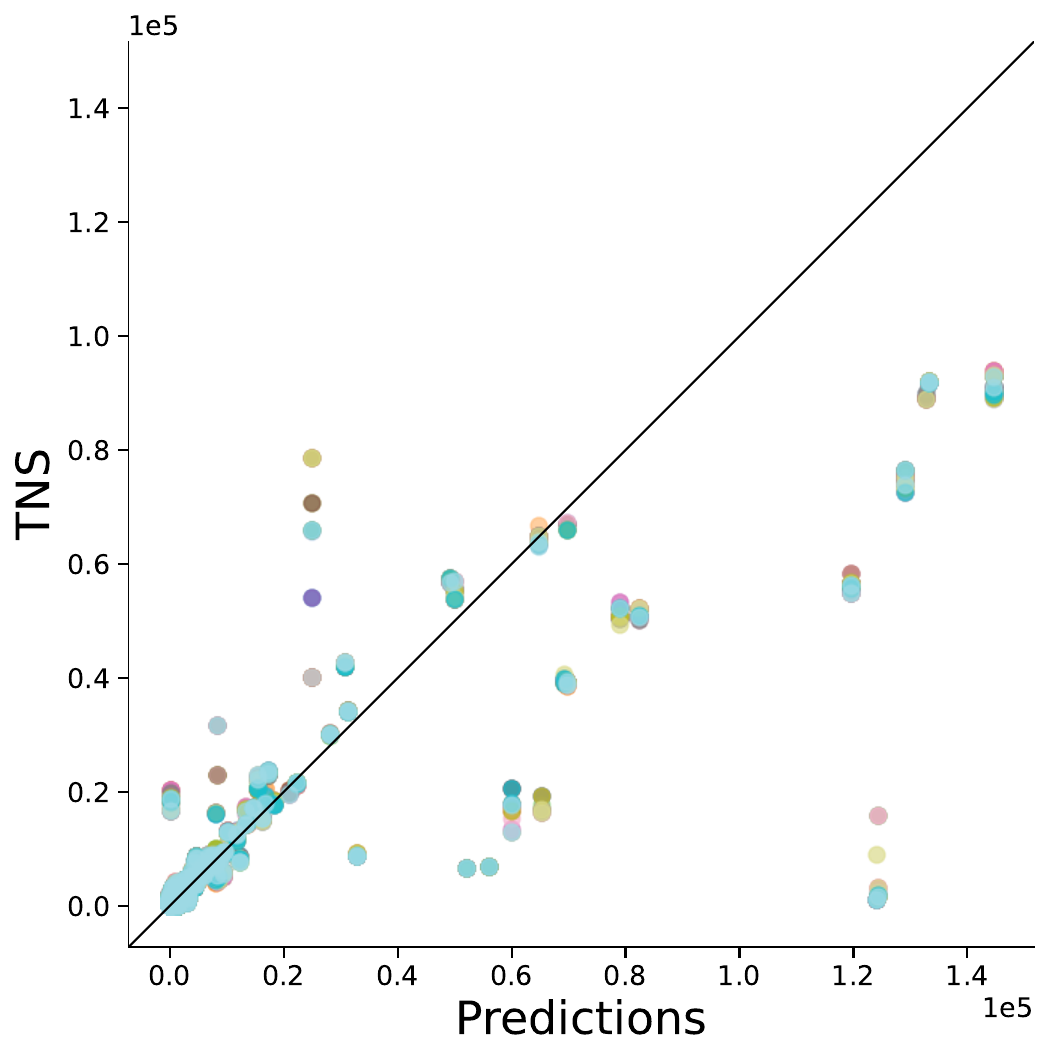} & \scat{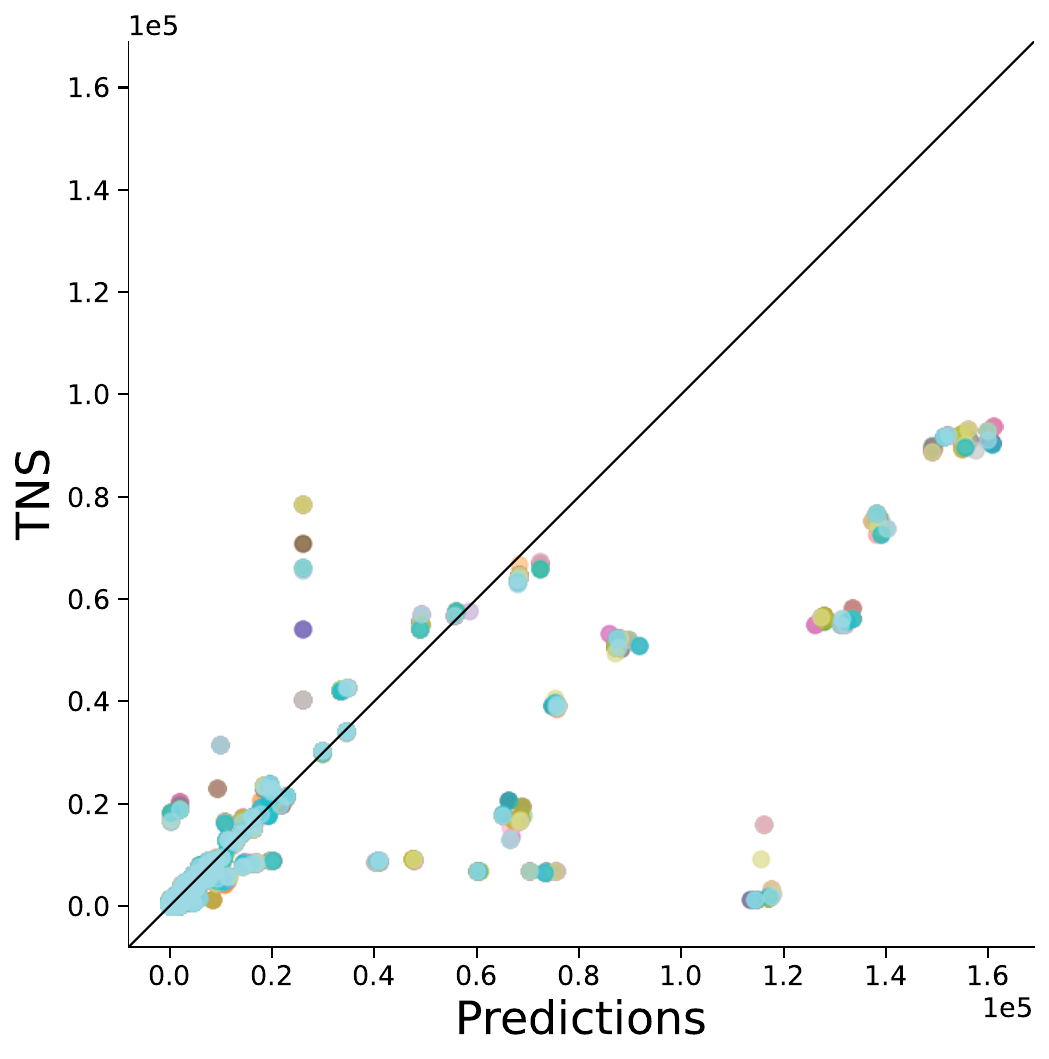} & \scat{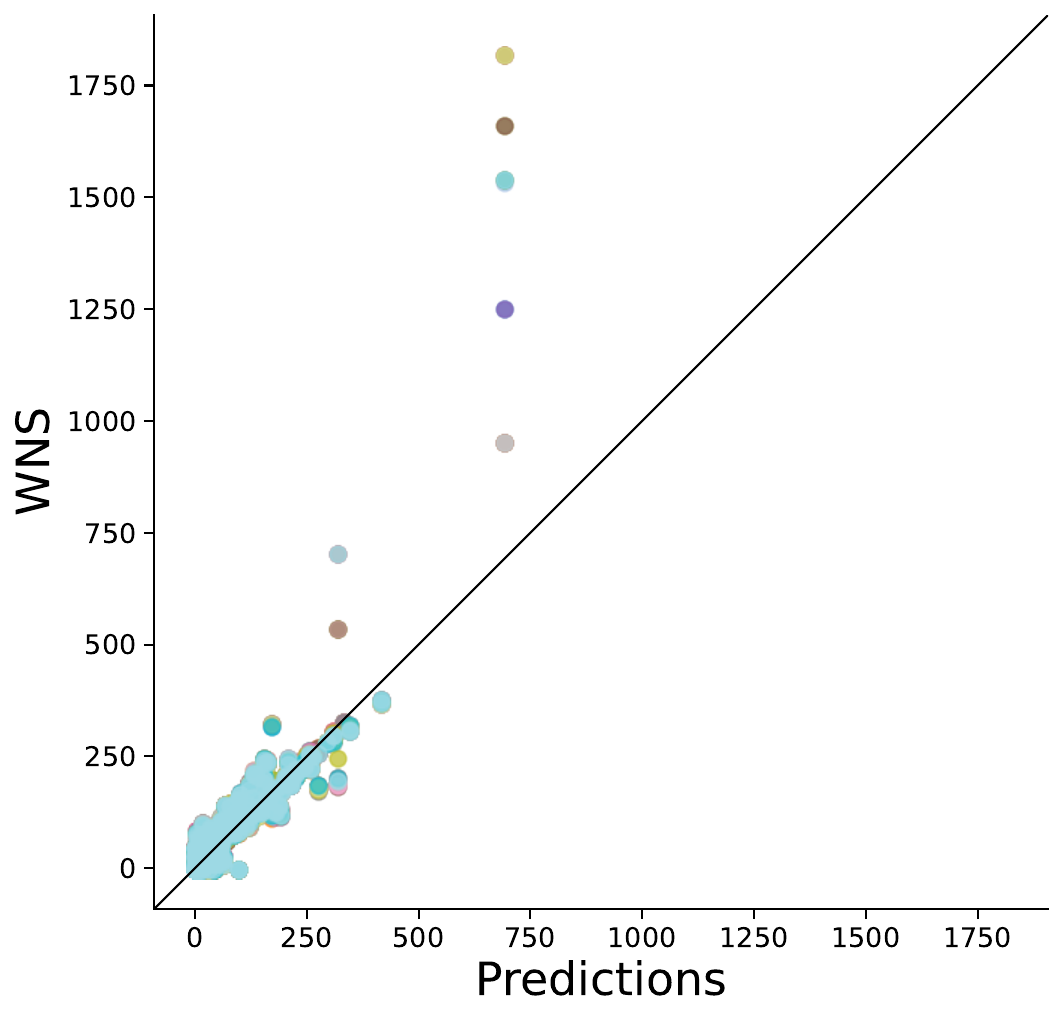} & \scat{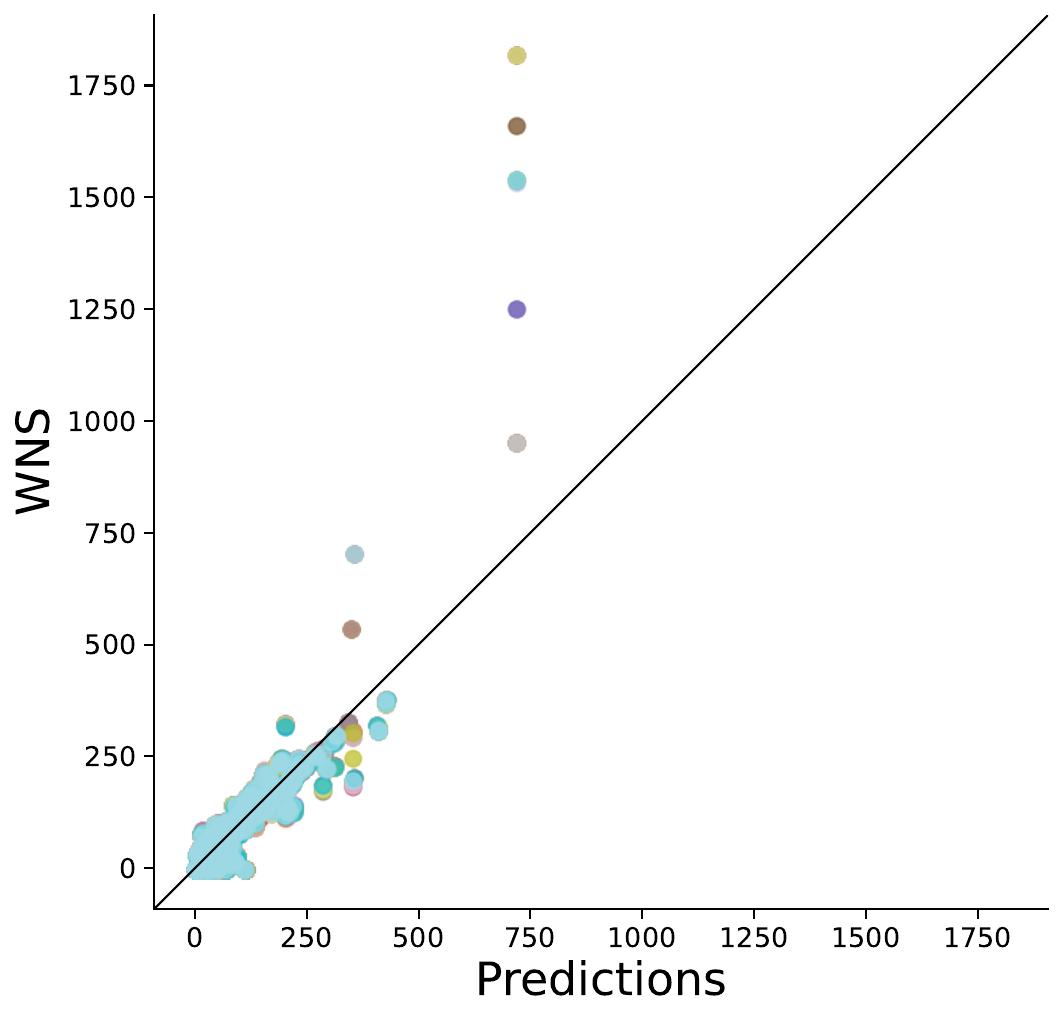} \\
\rowlab{OpenLLM-RTL} & \scat{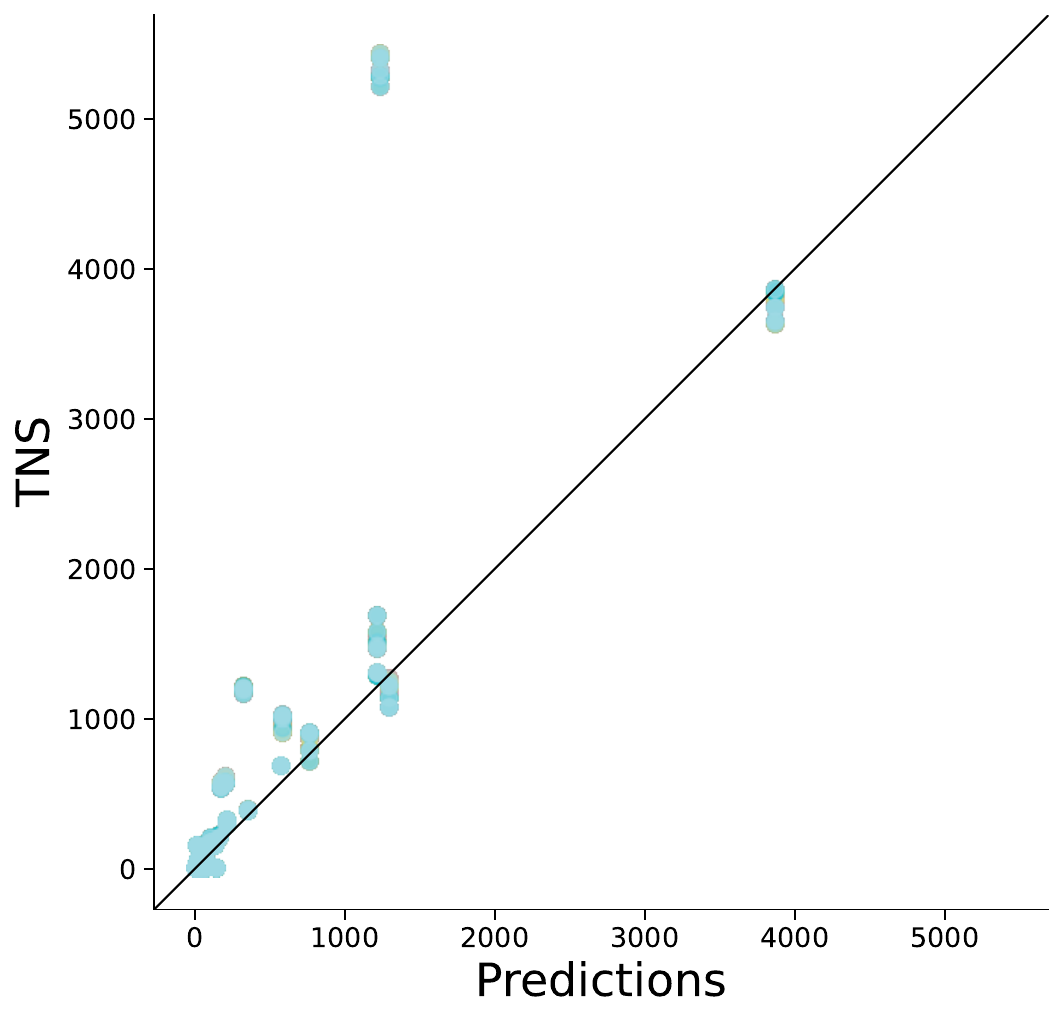} & \scat{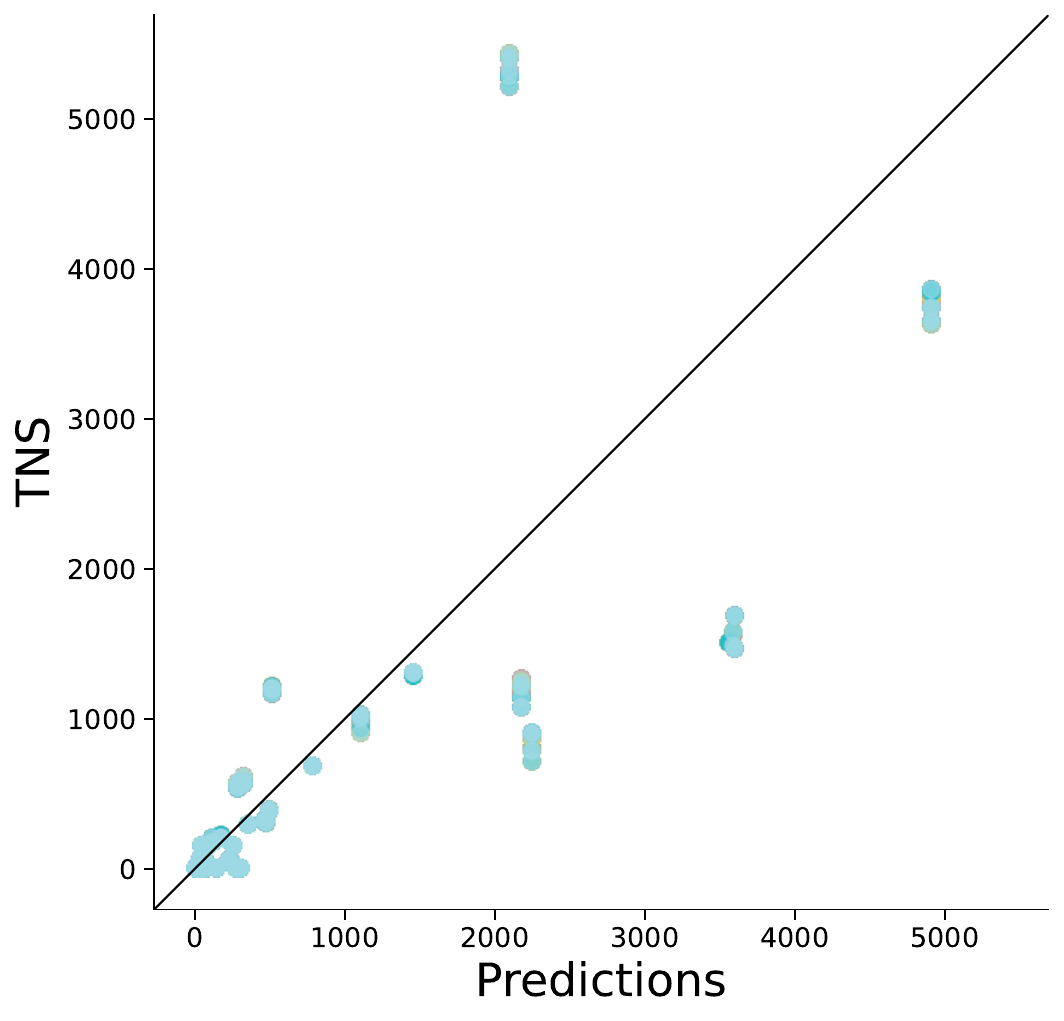} & \scat{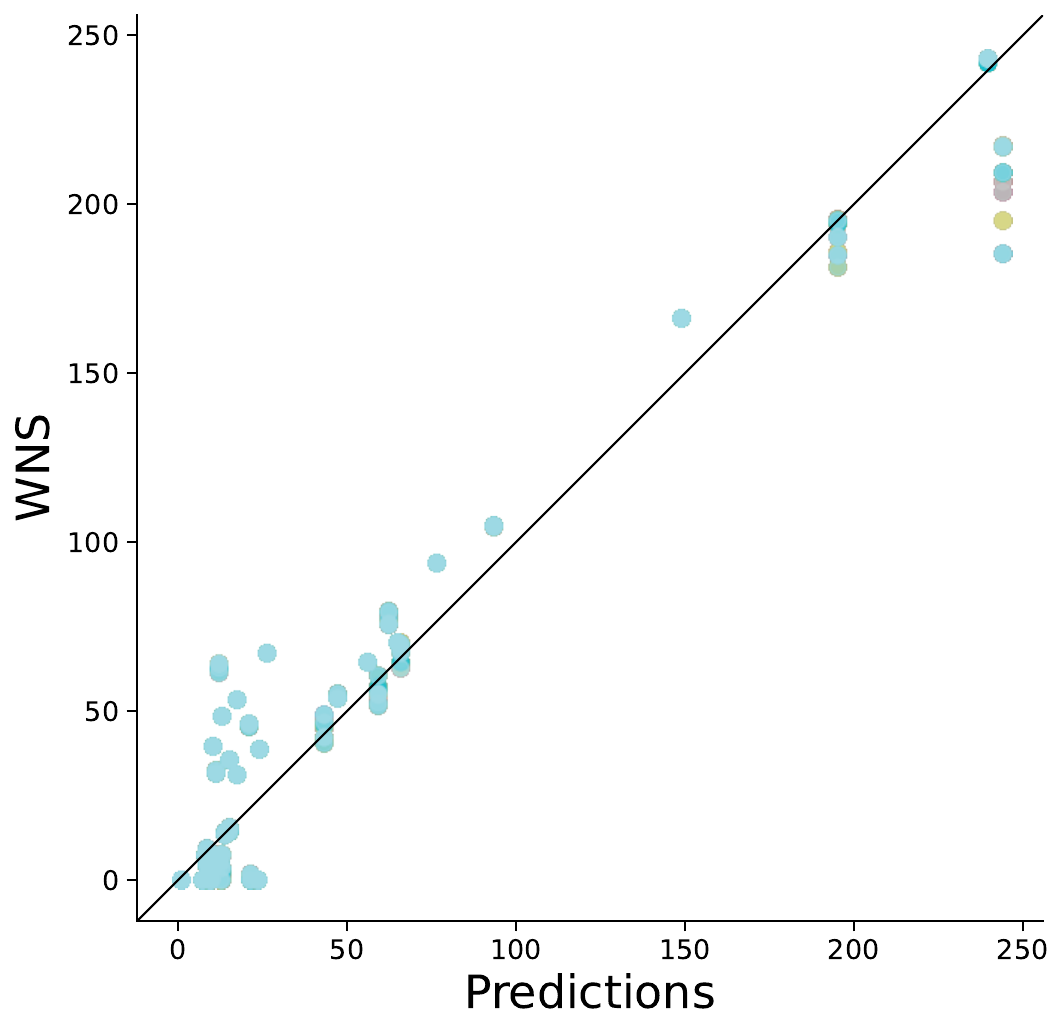} & \scat{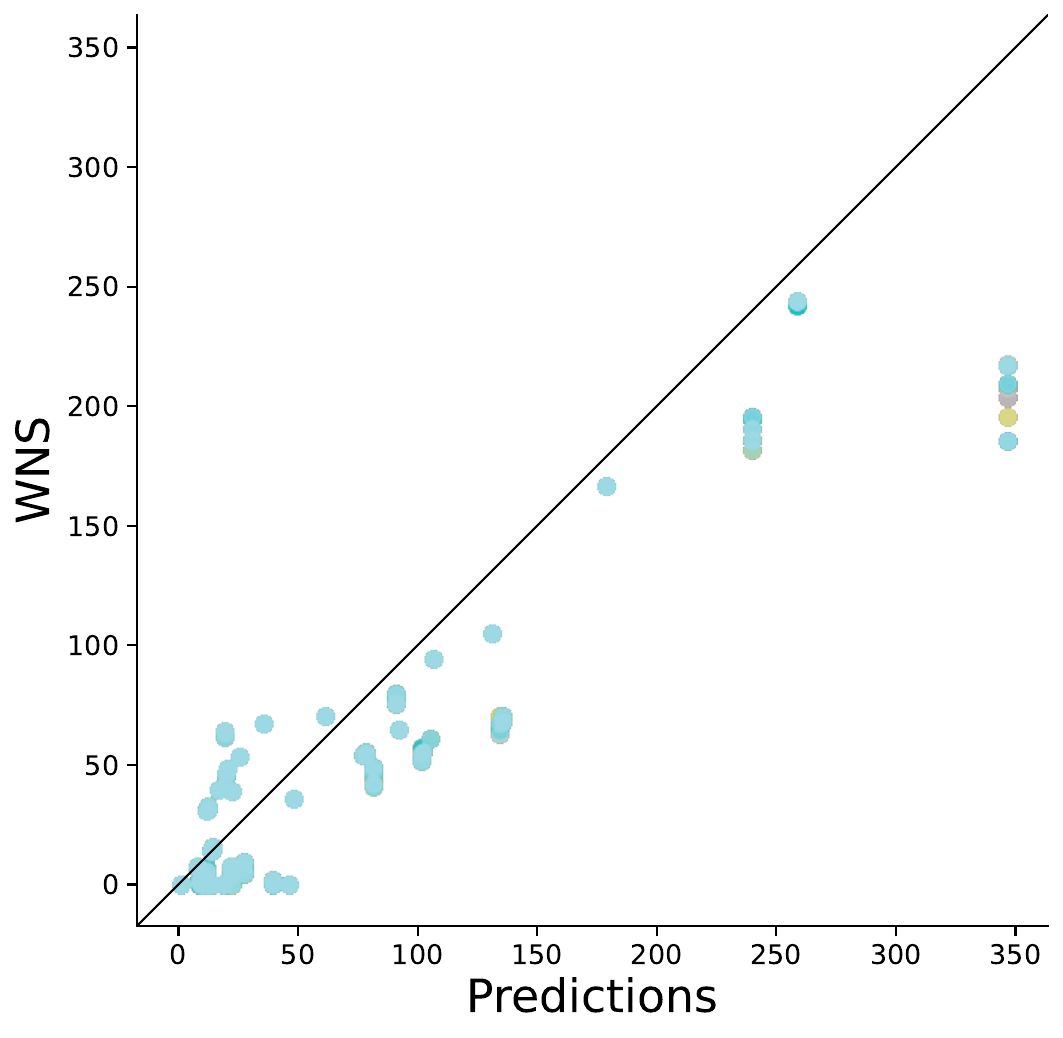} \\
\end{tabular}
\caption{Post-routing TNS and WNS (magnitudes, vertical axis) versus the STA estimate at the post-floorplan (PF) and post-placement (PP) stages (horizontal axis) on the held-out test set (top) and the OpenLLM-RTL benchmark (bottom). The diagonal marks agreement between the intermediate-stage and post-routing STA.}
\label{fig:scatter_stages}
\end{figure}

\begin{figure}[t]
\centering
\setlength{\tabcolsep}{1pt}
\begin{tabular}{cccc}
\small Held-out, TNS & \small Held-out, WNS & \small OpenLLM-RTL, TNS & \small OpenLLM-RTL, WNS \\
\scat{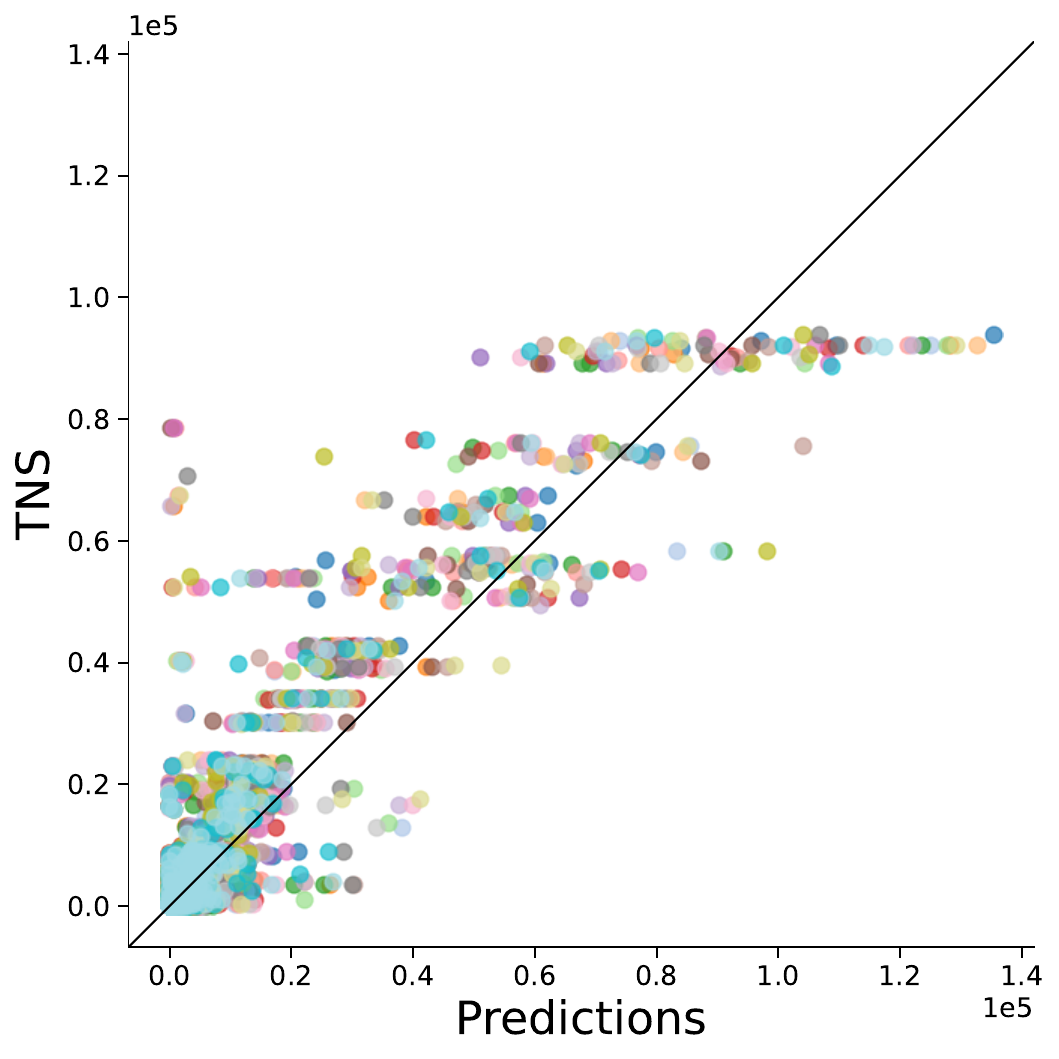} & \scat{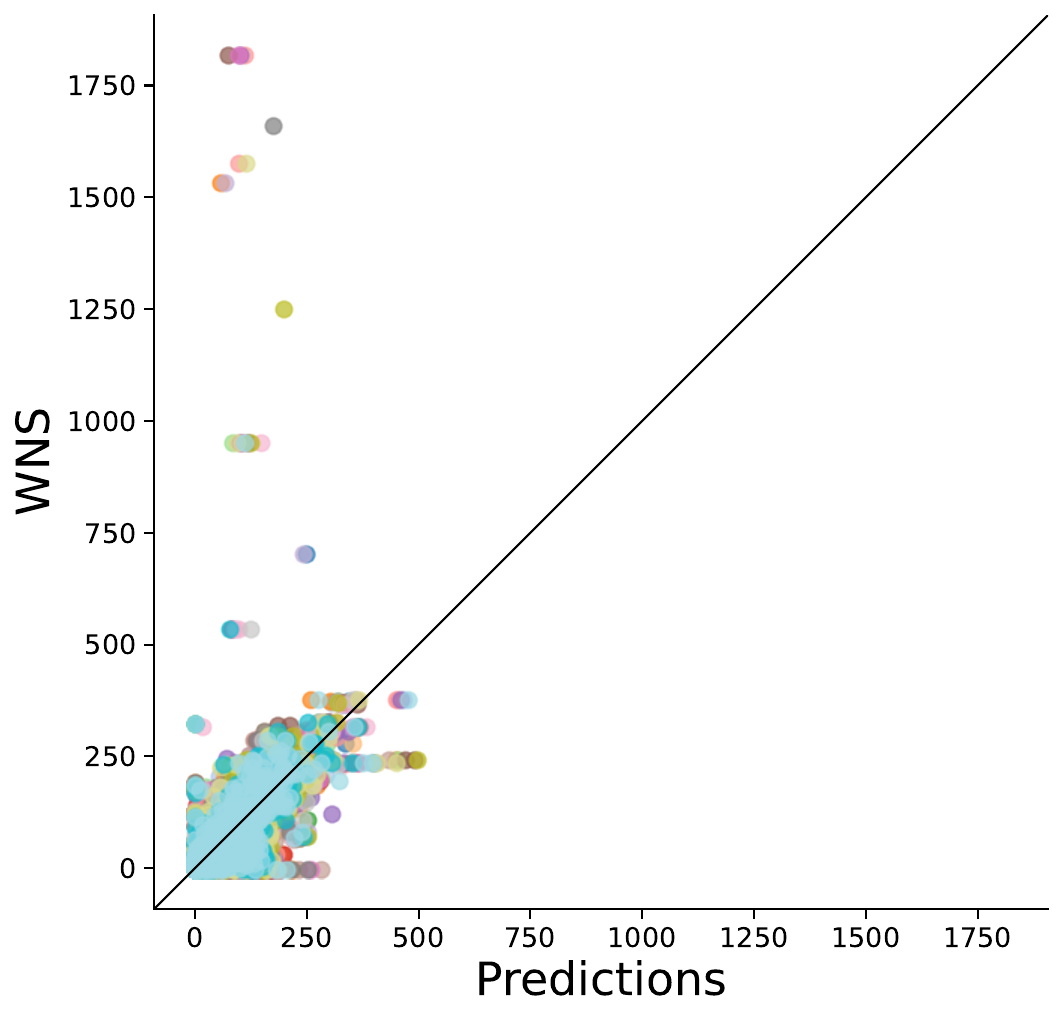} & \scat{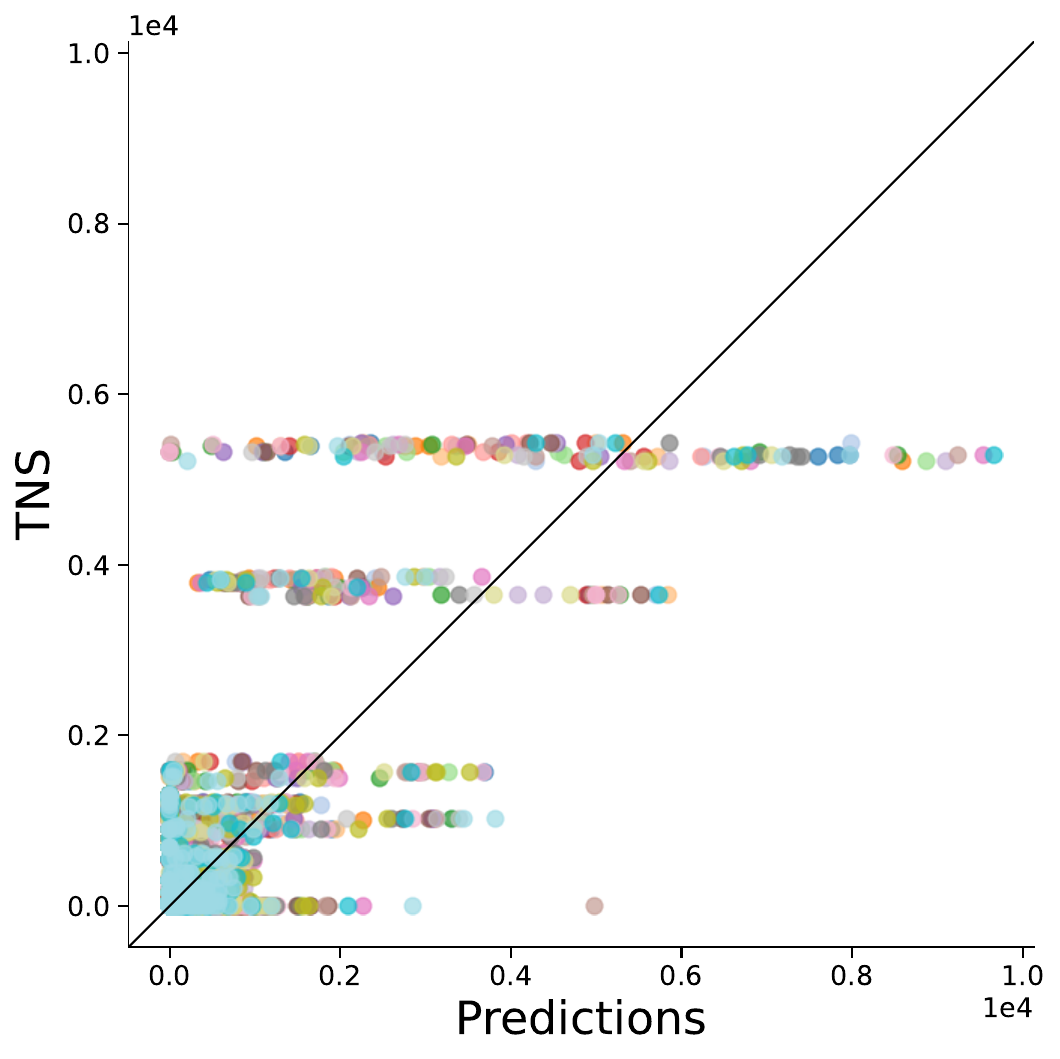} & \scat{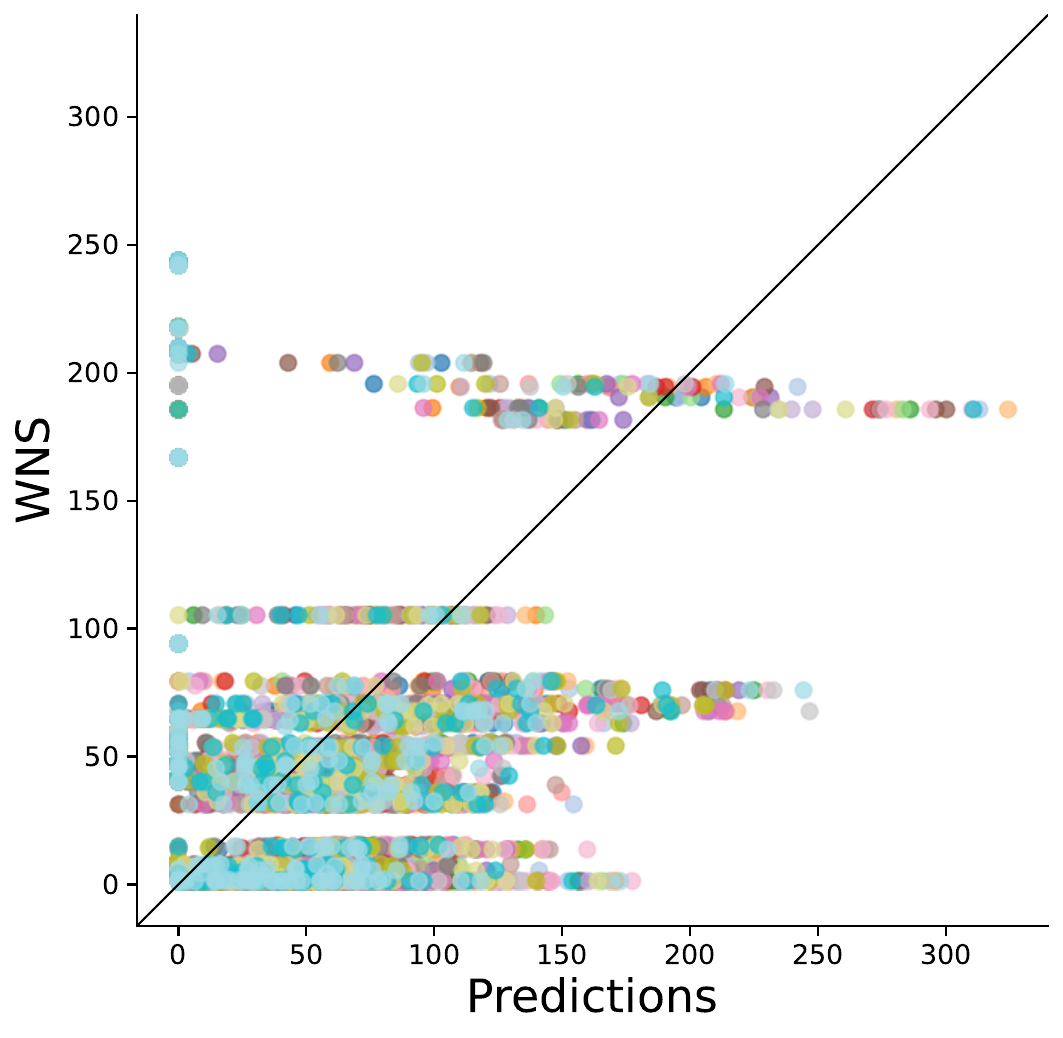} \\
\end{tabular}
\caption{Predicted versus post-routing TNS and WNS (magnitudes) for the No KD baseline, in the same format as Figure~\ref{fig:scatter}.}
\label{fig:scatter_nokd}
\end{figure}

Figure~\ref{fig:scatter_stages} compares the post-routing TNS and WNS with the STA estimates at the post-floorplan and post-placement stages, and Figure~\ref{fig:scatter_nokd} shows the predictions of the No KD baseline as a reference for Figure~\ref{fig:scatter}.

\textbf{Timing changes across design stages.} The intermediate-stage STA estimates show that the circuits with the largest post-routing WNS on the held-out set are underestimated at every earlier stage, although the estimates increase along the flow: from about 200 at post-synthesis (Figure~\ref{fig:scatter}) to about 700 at the PF and PP stages (Figure~\ref{fig:scatter_stages}), against a post-routing WNS of about 950 to 1{,}820. Their post-routing WNS also differs considerably between implementations that share the same PF and PP estimate.

\textbf{Sensitivity to implementation parameters.} The implementations of a design differ only in their OpenROAD parameters (core utilization and placement density) and are otherwise identical at the post-synthesis stage, yet these parameters largely determine the circuit's timing after the later stages. Post-synthesis STA does not use them, so it predicts the same value for every implementation of a design: its predictions form vertical columns in Figure~\ref{fig:scatter}, and the points from the different evaluation runs overlap, even where the post-routing timing differs substantially across implementations. STEP-KD receives these parameters as input features, and its predictions change with them, which appears as horizontal spread and as distinct colors from different runs. Taking the physical design parameters into account is part of STEP-KD's advantage over STA and over prior work that estimates timing before the physical design stages. On OpenLLM-RTL, where the post-routing timing of a design varies little across implementations, this sensitivity also shows up as spread of the predictions around the label, suggesting room to improve how the model uses these parameters. The No KD baseline (Figure~\ref{fig:scatter_nokd}), which receives the same input features, shows a much wider spread: on OpenLLM-RTL, for example, its TNS predictions for the same design range over several thousand, and several circuits with a large post-routing WNS are predicted to have almost none. Distillation from the later stages therefore makes the use of these parameters considerably more precise.

\section{Complete Ablation Results}
\label{app:ablations}
Table~\ref{tab:main-r2} reports the results of Table~\ref{tab:main} with $R^2$ and MAE, and Tables~\ref{tab:app-granularity} and~\ref{tab:app-depth} report all metrics for the ablations in Figures~\ref{fig:multigranularity_ablation} and~\ref{fig:depth}.

\input{tables/main_r2_arxiv}
\input{tables/heldout_app-granularity.tex}
\input{tables/heldout_app-depth.tex}

\section{Results on OpenLLM-RTL}
\label{app:olr}
OpenLLM-RTL main results are reported in Table~\ref{tab:main} (bottom) and Figures~\ref{fig:multigranularity_ablation} and~\ref{fig:depth}. Table~\ref{tab:intermediate-olr} mirrors Table~\ref{tab:intermediate}, and Tables~\ref{tab:app-granularity-olr} and~\ref{tab:app-depth-olr} give complete results for the ablations.

\input{tables/openllm_rtl_intermediate.tex}
\input{tables/openllm_rtl_app-granularity.tex}
\input{tables/openllm_rtl_app-depth.tex}

\end{document}

%% file: tables/main_combined.tex
\begin{table}[t]
\centering
\caption{Performance of methods taking post-synthesis inputs and predicting post-routing timing on the held-out test set (top) and the OpenLLM-RTL benchmark (bottom). Direct KD distills from a single teacher (stage in parentheses). STEP-KD w/o sequential KD aligns the student to independently trained PF, PP and PR teachers; STEP-KD uses teachers that were themselves distilled along the design ladder (PR$\rightarrow$PP$\rightarrow$PF). MedAE (median absolute error) is in the units of the slack labels. Best per dataset in bold.}
\label{tab:main}
\label{tab:main-olr}
\resizebox{\textwidth}{!}{%
\begin{tabular}{lcccccc}
\toprule
 & \multicolumn{3}{c}{TNS} & \multicolumn{3}{c}{WNS} \\
\cmidrule(lr){2-4}\cmidrule(lr){5-7}
Method & MedAE $\downarrow$ & WMAPE (\%) $\downarrow$ & RSE (\%) $\downarrow$ & MedAE $\downarrow$ & WMAPE (\%) $\downarrow$ & RSE (\%) $\downarrow$ \\
\midrule
\multicolumn{7}{l}{\textit{Held-out test set}} \\
\midrule
STA & $512.61 \pm 20.42$ & $74.84 \pm 0.60$ & $145.49 \pm 3.11$ & $23.69 \pm 0.56$ & $35.59 \pm 1.53$ & $46.52 \pm 14.72$ \\
No KD & $740.05 \pm 52.44$ & $38.75 \pm 2.20$ & $16.83 \pm 4.06$ & $25.82 \pm 1.73$ & $44.49 \pm 2.23$ & $62.14 \pm 14.24$ \\
\cmidrule(lr){1-7}
Direct KD (PR teacher) & $399.07 \pm 27.79$ & $24.05 \pm 1.51$ & $11.24 \pm 3.13$ & $15.45 \pm 0.99$ & $28.64 \pm 2.30$ & $45.16 \pm 18.84$ \\
Direct KD (PP teacher) & $396.84 \pm 31.08$ & $24.49 \pm 1.84$ & $10.95 \pm 3.71$ & $16.56 \pm 0.83$ & $29.14 \pm 2.12$ & $45.33 \pm 19.40$ \\
Direct KD (PF teacher) & $421.20 \pm 30.84$ & $23.65 \pm 1.54$ & $9.79 \pm 3.37$ & $18.23 \pm 0.97$ & $29.29 \pm 2.31$ & $43.03 \pm 20.25$ \\
\cmidrule(lr){1-7}
STEP-KD w/o sequential KD & $403.09 \pm 27.43$ & $30.18 \pm 1.79$ & $13.30 \pm 3.36$ & $16.35 \pm 0.59$ & $26.91 \pm 1.88$ & $44.96 \pm 17.65$ \\
STEP-KD & $\mathbf{302.27 \pm 18.37}$ & $\mathbf{19.78 \pm 1.54}$ & $\mathbf{7.85 \pm 2.95}$ & $\mathbf{11.09 \pm 0.41}$ & $\mathbf{21.39 \pm 1.97}$ & $\mathbf{41.09 \pm 17.81}$ \\
\midrule\midrule
\multicolumn{7}{l}{\textit{OpenLLM-RTL benchmark}} \\
\midrule
STA & $\mathbf{50.76 \pm 0.83}$ & $41.84 \pm 0.43$ & $43.42 \pm 0.50$ & $\mathbf{13.24 \pm 0.25}$ & $\mathbf{33.71 \pm 0.39}$ & $16.88 \pm 0.68$ \\
No KD & $188.01 \pm 17.06$ & $79.37 \pm 4.03$ & $52.07 \pm 9.63$ & $37.05 \pm 3.23$ & $99.34 \pm 3.75$ & $158.13 \pm 7.75$ \\
\cmidrule(lr){1-7}
Direct KD (PR teacher) & $73.55 \pm 6.72$ & $43.78 \pm 2.59$ & $25.08 \pm 4.51$ & $21.18 \pm 1.70$ & $54.62 \pm 2.20$ & $43.40 \pm 2.22$ \\
Direct KD (PP teacher) & $72.99 \pm 5.98$ & $49.84 \pm 3.14$ & $42.04 \pm 9.76$ & $16.51 \pm 1.32$ & $45.25 \pm 2.21$ & $32.44 \pm 2.89$ \\
Direct KD (PF teacher) & $101.47 \pm 7.10$ & $38.23 \pm 3.00$ & $14.86 \pm 4.50$ & $23.38 \pm 1.73$ & $50.30 \pm 1.67$ & $31.22 \pm 1.86$ \\
\cmidrule(lr){1-7}
STEP-KD w/o sequential KD & $76.74 \pm 5.08$ & $32.41 \pm 2.67$ & $11.93 \pm 4.43$ & $14.10 \pm 0.71$ & $35.58 \pm 0.61$ & $20.08 \pm 0.71$ \\
STEP-KD & $62.43 \pm 4.14$ & $\mathbf{27.43 \pm 1.41}$ & $\mathbf{6.49 \pm 1.23}$ & $15.98 \pm 0.69$ & $36.55 \pm 0.72$ & $\mathbf{16.52 \pm 0.64}$ \\
\bottomrule
\end{tabular}}
\end{table}

%% file: tables/heldout_intermediate.tex
\begin{table}[t]
\centering
\caption{STA, teacher models and the effect of sequential knowledge distillation (held-out test set). At each intermediate stage we compare STA, a model trained without distillation from later stages, and a model distilled along the design ladder. At PS both STEP-KD variants align to PF, PP and PR teachers one at a time and differ only in whether those teachers were themselves distilled; post-synthesis STA is reported in Table~\ref{tab:main}. MedAE (median absolute error) is in the units of the slack labels. Best within each stage in bold. $^\dagger$The PR teacher receives post-routing slack as an input feature, so its near-perfect score is expected and shown only for completeness.}
\label{tab:intermediate}
\label{tab:teachers}
\resizebox{\textwidth}{!}{%
\begin{tabular}{llcccccc}
\toprule
 & & \multicolumn{3}{c}{TNS} & \multicolumn{3}{c}{WNS} \\
\cmidrule(lr){3-5}\cmidrule(lr){6-8}
Stage & Predictor & MedAE $\downarrow$ & WMAPE (\%) $\downarrow$ & RSE (\%) $\downarrow$ & MedAE $\downarrow$ & WMAPE (\%) $\downarrow$ & RSE (\%) $\downarrow$ \\
\midrule
\multirow{2}{*}{PR} & STA & $0.00 \pm 0.00$ & $0.00 \pm 0.00$ & $0.00 \pm 0.00$ & $0.00 \pm 0.00$ & $0.00 \pm 0.00$ & $0.00 \pm 0.00$ \\
 & Teacher$^\dagger$ & $32.61 \pm 2.88$ & $1.74 \pm 0.15$ & $0.06 \pm 0.03$ & $1.12 \pm 0.08$ & $1.88 \pm 0.22$ & $0.23 \pm 0.21$ \\
\midrule
\multirow{3}{*}{PP} & STA & $378.28 \pm 16.03$ & $63.57 \pm 0.91$ & $105.06 \pm 3.82$ & $18.80 \pm 0.61$ & $29.71 \pm 1.06$ & $22.64 \pm 5.65$ \\
 & Independent (no KD) & $355.77 \pm 24.70$ & $19.61 \pm 1.52$ & $6.01 \pm 2.13$ & $\mathbf{12.11 \pm 0.73}$ & $\mathbf{21.22 \pm 1.53}$ & $20.09 \pm 9.96$ \\
 & KD from PR & $\mathbf{306.37 \pm 19.85}$ & $\mathbf{17.29 \pm 1.02}$ & $\mathbf{4.56 \pm 1.19}$ & $14.46 \pm 0.60$ & $22.78 \pm 1.08$ & $\mathbf{17.06 \pm 5.26}$ \\
\midrule
\multirow{3}{*}{PF} & STA & $424.67 \pm 16.87$ & $57.39 \pm 0.58$ & $85.85 \pm 2.42$ & $18.36 \pm 0.70$ & $30.29 \pm 0.91$ & $24.41 \pm 5.63$ \\
 & Independent (no KD) & $341.25 \pm 28.51$ & $20.66 \pm 1.72$ & $6.48 \pm 2.13$ & $15.57 \pm 0.78$ & $23.46 \pm 1.47$ & $\mathbf{20.74 \pm 9.82}$ \\
 & KD from PP (PR$\rightarrow$PP) & $\mathbf{273.94 \pm 19.38}$ & $\mathbf{17.43 \pm 1.62}$ & $\mathbf{5.38 \pm 2.19}$ & $\mathbf{12.48 \pm 0.59}$ & $\mathbf{20.58 \pm 1.90}$ & $20.80 \pm 9.64$ \\
\midrule
\multirow{2}{*}{PS} & STEP-KD w/o sequential KD & $403.09 \pm 27.43$ & $30.18 \pm 1.79$ & $13.30 \pm 3.36$ & $16.35 \pm 0.59$ & $26.91 \pm 1.88$ & $44.96 \pm 17.65$ \\
 & STEP-KD & $\mathbf{302.27 \pm 18.37}$ & $\mathbf{19.78 \pm 1.54}$ & $\mathbf{7.85 \pm 2.95}$ & $\mathbf{11.09 \pm 0.41}$ & $\mathbf{21.39 \pm 1.97}$ & $\mathbf{41.09 \pm 17.81}$ \\
\bottomrule
\end{tabular}}
\end{table}

%% file: tables/slack_bias.tex
\begin{table}[t]
\centering
\caption{Node-level input of the timing predictor for the post-synthesis model trained with Direct KD from the PR teacher (held-out test set). Row (iii) is the configuration used throughout the paper and corresponds to Direct KD (PR teacher) in Table~\ref{tab:main}. MedAE is in the units of the slack labels. Best in bold.}
\label{tab:slack_bias}
\resizebox{\textwidth}{!}{%
\begin{tabular}{lcccccc}
\toprule
 & \multicolumn{3}{c}{TNS} & \multicolumn{3}{c}{WNS} \\
\cmidrule(lr){2-4}\cmidrule(lr){5-7}
Node-level representation & MedAE $\downarrow$ & WMAPE (\%) $\downarrow$ & RSE (\%) $\downarrow$ & MedAE $\downarrow$ & WMAPE (\%) $\downarrow$ & RSE (\%) $\downarrow$ \\
\midrule
(i) GNN output $\mathbf{z}_r$ & $2201.40 \pm 132.98$ & $78.78 \pm 2.45$ & $60.46 \pm 5.65$ & $55.85 \pm 1.66$ & $75.76 \pm 1.90$ & $103.59 \pm 5.26$ \\
(ii) Slack embedding $p(\text{slack}(r))$ & $495.26 \pm 33.28$ & $24.97 \pm 2.01$ & $11.48 \pm 3.52$ & $15.83 \pm 0.94$ & $29.38 \pm 2.16$ & $\mathbf{43.00 \pm 18.45}$ \\
(iii) $\mathbf{z}_r + p(\text{slack}(r))$ (ours) & $\mathbf{399.07 \pm 27.79}$ & $\mathbf{24.05 \pm 1.51}$ & $\mathbf{11.24 \pm 3.13}$ & $\mathbf{15.45 \pm 0.99}$ & $\mathbf{28.64 \pm 2.30}$ & $45.16 \pm 18.84$ \\
\bottomrule
\end{tabular}}
\end{table}

%% file: tables/main_r2_arxiv.tex
\begin{table}[t]
\tiny
\centering
\caption{Results of Table~\ref{tab:main} reported with $R^2$ and the mean absolute error (MAE, in the units of the slack labels) on the held-out test set (top) and the OpenLLM-RTL benchmark (bottom). Best per dataset in bold.}
\label{tab:main-r2}
\resizebox{0.8\textwidth}{!}{%  <-- Changed from \textwidth to 0.75\textwidth
\begin{tabular}{lcccc}
\toprule
 & \multicolumn{2}{c}{TNS} & \multicolumn{2}{c}{WNS} \\
\cmidrule(lr){2-3}\cmidrule(lr){4-5}
Method & R$^2$ (\%) $\uparrow$ & MAE $\downarrow$ & R$^2$ (\%) $\uparrow$ & MAE $\downarrow$ \\
\midrule
\multicolumn{5}{l}{\textit{Held-out test set}} \\
\midrule
STA & $-45.49 \pm 3.11$ & $5072.45 \pm 106.68$ & $53.48 \pm 14.72$ & $32.25 \pm 2.32$ \\
No KD & $83.17 \pm 4.06$ & $2604.43 \pm 180.34$ & $37.86 \pm 14.24$ & $39.81 \pm 3.06$ \\
\cmidrule(lr){1-5}
Direct KD (PR teacher) & $88.76 \pm 3.13$ & $1619.63 \pm 126.85$ & $54.84 \pm 18.84$ & $25.74 \pm 2.80$ \\
Direct KD (PP teacher) & $89.05 \pm 3.71$ & $1649.69 \pm 151.67$ & $54.67 \pm 19.40$ & $26.17 \pm 2.64$ \\
Direct KD (PF teacher) & $90.21 \pm 3.37$ & $1592.88 \pm 130.81$ & $56.97 \pm 20.25$ & $26.32 \pm 2.80$ \\
\cmidrule(lr){1-5}
STEP-KD w/o sequential KD & $86.70 \pm 3.36$ & $2028.72 \pm 148.66$ & $55.04 \pm 17.65$ & $24.04 \pm 2.28$ \\
STEP-KD & $\mathbf{92.15 \pm 2.95}$ & $\mathbf{1330.00 \pm 125.66}$ & $\mathbf{58.91 \pm 17.81}$ & $\mathbf{19.12 \pm 2.24}$ \\
\midrule\midrule
\multicolumn{5}{l}{\textit{OpenLLM-RTL benchmark}} \\
\midrule
STA & $56.58 \pm 0.50$ & $221.75 \pm 2.13$ & $83.12 \pm 0.68$ & $\mathbf{17.32 \pm 0.15}$ \\
No KD & $47.93 \pm 9.63$ & $420.68 \pm 21.49$ & $-58.13 \pm 7.75$ & $51.03 \pm 1.91$ \\
\cmidrule(lr){1-5}
Direct KD (PR teacher) & $74.92 \pm 4.51$ & $232.25 \pm 13.96$ & $56.60 \pm 2.22$ & $28.07 \pm 1.11$ \\
Direct KD (PP teacher) & $57.96 \pm 9.76$ & $264.40 \pm 16.85$ & $67.56 \pm 2.89$ & $23.25 \pm 1.11$ \\
Direct KD (PF teacher) & $85.14 \pm 4.50$ & $202.79 \pm 15.89$ & $68.78 \pm 1.86$ & $25.85 \pm 0.83$ \\
\cmidrule(lr){1-5}
STEP-KD w/o sequential KD & $88.07 \pm 4.43$ & $172.07 \pm 14.10$ & $79.92 \pm 0.71$ & $18.29 \pm 0.31$ \\
STEP-KD & $\mathbf{93.51 \pm 1.23}$ & $\mathbf{145.60 \pm 7.50}$ & $\mathbf{83.48 \pm 0.64}$ & $18.78 \pm 0.35$ \\
\bottomrule
\end{tabular}}
\end{table}

%% file: tables/heldout_app-granularity.tex
\begin{table}[t]
\centering
\caption{Complete results for the multi-granularity ablation on the held-out test set (Figure~\ref{fig:multigranularity_ablation}). Each variant distills from the same PF, PP and PR teachers as STEP-KD but aligns only the listed granularity; \emph{All} is the full STEP-KD. No KD is included for reference. MAE is in the units of the slack labels. Best in bold.}
\label{tab:app-granularity}
\resizebox{\textwidth}{!}{%
\begin{tabular}{lcccccccc}
\toprule
 & \multicolumn{4}{c}{TNS} & \multicolumn{4}{c}{WNS} \\
\cmidrule(lr){2-5}\cmidrule(lr){6-9}
KD granularity & R$^2$ (\%) $\uparrow$ & WMAPE (\%) $\downarrow$ & RSE (\%) $\downarrow$ & MAE $\downarrow$ & R$^2$ (\%) $\uparrow$ & WMAPE (\%) $\downarrow$ & RSE (\%) $\downarrow$ & MAE $\downarrow$ \\
\midrule
No KD (reference) & $83.17 \pm 4.06$ & $38.75 \pm 2.20$ & $16.83 \pm 4.06$ & $2604.43 \pm 180.34$ & $37.86 \pm 14.24$ & $44.49 \pm 2.23$ & $62.14 \pm 14.24$ & $39.81 \pm 3.06$ \\
Graph only & $90.74 \pm 2.78$ & $24.07 \pm 1.27$ & $9.26 \pm 2.78$ & $1617.55 \pm 110.02$ & $56.93 \pm 16.72$ & $26.53 \pm 1.82$ & $43.07 \pm 16.72$ & $23.70 \pm 2.21$ \\
Subgraph only & $49.72 \pm 3.98$ & $56.36 \pm 1.58$ & $50.28 \pm 3.98$ & $3786.66 \pm 155.30$ & $52.70 \pm 14.19$ & $38.69 \pm 1.54$ & $47.30 \pm 14.19$ & $34.54 \pm 2.22$ \\
Node only & $89.84 \pm 3.22$ & $26.95 \pm 1.92$ & $10.16 \pm 3.22$ & $1812.13 \pm 158.15$ & $49.84 \pm 16.68$ & $30.62 \pm 1.82$ & $50.16 \pm 16.68$ & $27.35 \pm 2.29$ \\
All (STEP-KD) & $\mathbf{92.15 \pm 2.95}$ & $\mathbf{19.78 \pm 1.54}$ & $\mathbf{7.85 \pm 2.95}$ & $\mathbf{1330.00 \pm 125.66}$ & $\mathbf{58.91 \pm 17.81}$ & $\mathbf{21.39 \pm 1.97}$ & $\mathbf{41.09 \pm 17.81}$ & $\mathbf{19.12 \pm 2.24}$ \\
\bottomrule
\end{tabular}}
\end{table}

%% file: tables/heldout_app-depth.tex
\begin{table}[t]
\centering
\caption{Complete results for the projection-depth ablation of Direct KD from the PR teacher on the held-out test set (Figure~\ref{fig:depth}). The default Direct KD (PR teacher) in Table~\ref{tab:main} uses 3-layer projections. STEP-KD, whose PS model composes three 3-layer projections, is included for reference. Best in bold.}
\label{tab:app-depth}
\resizebox{\textwidth}{!}{%
\begin{tabular}{lcccccccc}
\toprule
 & \multicolumn{4}{c}{TNS} & \multicolumn{4}{c}{WNS} \\
\cmidrule(lr){2-5}\cmidrule(lr){6-9}
Projection layers & R$^2$ (\%) $\uparrow$ & WMAPE (\%) $\downarrow$ & RSE (\%) $\downarrow$ & MAE $\downarrow$ & R$^2$ (\%) $\uparrow$ & WMAPE (\%) $\downarrow$ & RSE (\%) $\downarrow$ & MAE $\downarrow$ \\
\midrule
4 & $88.45 \pm 2.89$ & $24.76 \pm 1.22$ & $11.55 \pm 2.89$ & $1673.53 \pm 108.72$ & $51.62 \pm 16.35$ & $26.74 \pm 2.01$ & $48.38 \pm 16.35$ & $24.19 \pm 2.45$ \\
6 & $88.46 \pm 4.16$ & $25.47 \pm 1.78$ & $11.54 \pm 4.16$ & $1730.35 \pm 159.76$ & $54.50 \pm 22.18$ & $27.34 \pm 2.76$ & $45.50 \pm 22.18$ & $24.82 \pm 3.45$ \\
8 & $88.10 \pm 3.84$ & $23.81 \pm 1.86$ & $11.90 \pm 3.84$ & $1625.50 \pm 158.01$ & $43.20 \pm 18.49$ & $31.93 \pm 2.42$ & $56.80 \pm 18.49$ & $29.32 \pm 3.18$ \\
10 & $90.42 \pm 2.91$ & $20.96 \pm 1.20$ & $9.58 \pm 2.91$ & $1423.49 \pm 99.74$ & $48.40 \pm 15.69$ & $25.91 \pm 1.81$ & $51.60 \pm 15.69$ & $23.61 \pm 2.18$ \\
STEP-KD ($3 \times 3$ layers) & $\mathbf{92.15 \pm 2.95}$ & $\mathbf{19.78 \pm 1.54}$ & $\mathbf{7.85 \pm 2.95}$ & $\mathbf{1330.00 \pm 125.66}$ & $\mathbf{58.91 \pm 17.81}$ & $\mathbf{21.39 \pm 1.97}$ & $\mathbf{41.09 \pm 17.81}$ & $\mathbf{19.12 \pm 2.24}$ \\
\bottomrule
\end{tabular}}
\end{table}

%% file: tables/openllm_rtl_intermediate.tex
\begin{table}[t]
\centering
\caption{STA, teacher models and the effect of sequential knowledge distillation (OpenLLM-RTL benchmark). At each intermediate stage we compare STA, a model trained without distillation from later stages, and a model distilled along the design ladder. At PS both STEP-KD variants align to PF, PP and PR teachers one at a time and differ only in whether those teachers were themselves distilled; post-synthesis STA is reported in Table~\ref{tab:main-olr}. MedAE (median absolute error) is in the units of the slack labels. Best within each stage in bold. $^\dagger$The PR teacher receives post-routing slack as an input feature, so its near-perfect score is expected and shown only for completeness.}
\label{tab:intermediate-olr}
\label{tab:teachers-olr}
\resizebox{\textwidth}{!}{%
\begin{tabular}{llcccccc}
\toprule
 & & \multicolumn{3}{c}{TNS} & \multicolumn{3}{c}{WNS} \\
\cmidrule(lr){3-5}\cmidrule(lr){6-8}
Stage & Predictor & MedAE $\downarrow$ & WMAPE (\%) $\downarrow$ & RSE (\%) $\downarrow$ & MedAE $\downarrow$ & WMAPE (\%) $\downarrow$ & RSE (\%) $\downarrow$ \\
\midrule
\multirow{2}{*}{PR} & STA & $0.00 \pm 0.00$ & $0.04 \pm 0.00$ & $0.00 \pm 0.00$ & $0.00 \pm 0.00$ & $0.39 \pm 0.02$ & $0.01 \pm 0.00$ \\
 & Teacher$^\dagger$ & $7.82 \pm 0.76$ & $4.11 \pm 0.32$ & $0.16 \pm 0.04$ & $1.30 \pm 0.15$ & $4.61 \pm 0.20$ & $0.56 \pm 0.04$ \\
\midrule
\multirow{3}{*}{PP} & STA & $84.52 \pm 7.03$ & $61.25 \pm 0.53$ & $47.19 \pm 0.67$ & $18.89 \pm 0.17$ & $48.28 \pm 0.45$ & $34.23 \pm 1.27$ \\
 & Independent (no KD) & $\mathbf{46.65 \pm 2.91}$ & $\mathbf{29.55 \pm 1.17}$ & $\mathbf{9.86 \pm 0.84}$ & $\mathbf{13.00 \pm 0.77}$ & $\mathbf{32.59 \pm 0.85}$ & $\mathbf{14.13 \pm 0.89}$ \\
 & KD from PR & $91.96 \pm 4.35$ & $40.77 \pm 1.32$ & $16.27 \pm 1.60$ & $18.68 \pm 0.65$ & $44.54 \pm 1.08$ & $23.47 \pm 1.08$ \\
\midrule
\multirow{3}{*}{PF} & STA & $\mathbf{51.20 \pm 0.43}$ & $40.18 \pm 0.33$ & $43.34 \pm 0.48$ & $\mathbf{9.60 \pm 0.19}$ & $\mathbf{27.06 \pm 0.35}$ & $\mathbf{10.28 \pm 0.35}$ \\
 & Independent (no KD) & $87.40 \pm 6.52$ & $48.94 \pm 1.91$ & $26.31 \pm 3.53$ & $22.24 \pm 1.40$ & $56.11 \pm 2.34$ & $42.83 \pm 3.30$ \\
 & KD from PP (PR$\rightarrow$PP) & $59.96 \pm 4.45$ & $\mathbf{27.49 \pm 1.81}$ & $\mathbf{13.30 \pm 3.25}$ & $17.57 \pm 0.99$ & $35.65 \pm 0.84$ & $13.02 \pm 0.57$ \\
\midrule
\multirow{2}{*}{PS} & STEP-KD w/o sequential KD & $76.74 \pm 5.08$ & $32.41 \pm 2.67$ & $11.93 \pm 4.43$ & $\mathbf{14.10 \pm 0.71}$ & $\mathbf{35.58 \pm 0.61}$ & $20.08 \pm 0.71$ \\
 & STEP-KD & $\mathbf{62.43 \pm 4.14}$ & $\mathbf{27.43 \pm 1.41}$ & $\mathbf{6.49 \pm 1.23}$ & $15.98 \pm 0.69$ & $36.55 \pm 0.72$ & $\mathbf{16.52 \pm 0.64}$ \\
\bottomrule
\end{tabular}}
\end{table}

%% file: tables/openllm_rtl_app-granularity.tex
\begin{table}[t]
\centering
\caption{Complete results for the multi-granularity ablation on the OpenLLM-RTL benchmark (Figure~\ref{fig:multigranularity_ablation}). Each variant distills from the same PF, PP and PR teachers as STEP-KD but aligns only the listed granularity; \emph{All} is the full STEP-KD. No KD is included for reference. MAE is in the units of the slack labels. Best in bold.}
\label{tab:app-granularity-olr}
\resizebox{\textwidth}{!}{%
\begin{tabular}{lcccccccc}
\toprule
 & \multicolumn{4}{c}{TNS} & \multicolumn{4}{c}{WNS} \\
\cmidrule(lr){2-5}\cmidrule(lr){6-9}
KD granularity & R$^2$ (\%) $\uparrow$ & WMAPE (\%) $\downarrow$ & RSE (\%) $\downarrow$ & MAE $\downarrow$ & R$^2$ (\%) $\uparrow$ & WMAPE (\%) $\downarrow$ & RSE (\%) $\downarrow$ & MAE $\downarrow$ \\
\midrule
No KD (reference) & $47.93 \pm 9.63$ & $79.37 \pm 4.03$ & $52.07 \pm 9.63$ & $420.68 \pm 21.49$ & $-58.13 \pm 7.75$ & $99.34 \pm 3.75$ & $158.13 \pm 7.75$ & $51.03 \pm 1.91$ \\
Graph only & $87.69 \pm 1.72$ & $37.60 \pm 1.76$ & $12.31 \pm 1.72$ & $199.62 \pm 9.28$ & $59.42 \pm 1.88$ & $51.50 \pm 1.32$ & $40.58 \pm 1.88$ & $26.47 \pm 0.66$ \\
Subgraph only & $72.15 \pm 2.78$ & $65.89 \pm 1.94$ & $27.85 \pm 2.78$ & $349.78 \pm 9.78$ & $42.28 \pm 2.67$ & $74.73 \pm 1.53$ & $57.72 \pm 2.67$ & $38.41 \pm 0.73$ \\
Node only & $87.43 \pm 1.09$ & $46.17 \pm 1.77$ & $12.57 \pm 1.09$ & $245.13 \pm 9.72$ & $37.91 \pm 2.50$ & $60.66 \pm 1.30$ & $62.09 \pm 2.50$ & $31.18 \pm 0.69$ \\
All (STEP-KD) & $\mathbf{93.51 \pm 1.23}$ & $\mathbf{27.43 \pm 1.41}$ & $\mathbf{6.49 \pm 1.23}$ & $\mathbf{145.60 \pm 7.50}$ & $\mathbf{83.48 \pm 0.64}$ & $\mathbf{36.55 \pm 0.72}$ & $\mathbf{16.52 \pm 0.64}$ & $\mathbf{18.78 \pm 0.35}$ \\
\bottomrule
\end{tabular}}
\end{table}

%% file: tables/openllm_rtl_app-depth.tex
\begin{table}[t]
\centering
\caption{Complete results for the projection-depth ablation of Direct KD from the PR teacher on the OpenLLM-RTL benchmark (Figure~\ref{fig:depth}). The default Direct KD (PR teacher) in Table~\ref{tab:main} uses 3-layer projections. STEP-KD, whose PS model composes three 3-layer projections, is included for reference. Best in bold.}
\label{tab:app-depth-olr}
\resizebox{\textwidth}{!}{%
\begin{tabular}{lcccccccc}
\toprule
 & \multicolumn{4}{c}{TNS} & \multicolumn{4}{c}{WNS} \\
\cmidrule(lr){2-5}\cmidrule(lr){6-9}
Projection layers & R$^2$ (\%) $\uparrow$ & WMAPE (\%) $\downarrow$ & RSE (\%) $\downarrow$ & MAE $\downarrow$ & R$^2$ (\%) $\uparrow$ & WMAPE (\%) $\downarrow$ & RSE (\%) $\downarrow$ & MAE $\downarrow$ \\
\midrule
4 & $88.02 \pm 2.49$ & $34.08 \pm 2.42$ & $11.98 \pm 2.49$ & $180.95 \pm 12.68$ & $54.56 \pm 2.49$ & $53.96 \pm 1.61$ & $45.44 \pm 2.49$ & $27.72 \pm 0.83$ \\
6 & $52.43 \pm 5.18$ & $51.66 \pm 1.56$ & $47.57 \pm 5.18$ & $274.15 \pm 8.13$ & $57.36 \pm 1.42$ & $53.70 \pm 1.16$ & $42.64 \pm 1.42$ & $27.61 \pm 0.60$ \\
8 & $38.41 \pm 8.33$ & $60.24 \pm 2.58$ & $61.59 \pm 8.33$ & $320.05 \pm 13.70$ & $39.30 \pm 3.16$ & $67.33 \pm 1.63$ & $60.70 \pm 3.16$ & $34.62 \pm 0.87$ \\
10 & $91.42 \pm 0.52$ & $32.64 \pm 0.64$ & $8.58 \pm 0.52$ & $173.11 \pm 3.57$ & $68.97 \pm 1.03$ & $41.95 \pm 0.72$ & $31.03 \pm 1.03$ & $21.56 \pm 0.33$ \\
STEP-KD ($3 \times 3$ layers) & $\mathbf{93.51 \pm 1.23}$ & $\mathbf{27.43 \pm 1.41}$ & $\mathbf{6.49 \pm 1.23}$ & $\mathbf{145.60 \pm 7.50}$ & $\mathbf{83.48 \pm 0.64}$ & $\mathbf{36.55 \pm 0.72}$ & $\mathbf{16.52 \pm 0.64}$ & $\mathbf{18.78 \pm 0.35}$ \\
\bottomrule
\end{tabular}}
\end{table}